\documentclass[10pt,twocolumn,letterpaper]{article}

\usepackage[pagenumbers]{cvpr}              

\usepackage{url}            
\usepackage{booktabs}       
\usepackage{amsfonts}       
\usepackage{nicefrac}       
\usepackage{microtype}      
\usepackage{graphicx}
\usepackage{amsmath}
\usepackage{amssymb}
\usepackage{array}
\usepackage{pifont}
\usepackage{capt-of,etoolbox}
\usepackage{multirow}
\usepackage{hhline}
\usepackage{bbding}
\usepackage{algorithm}
\usepackage{algorithmic}
\usepackage{boldline}
\usepackage{braket}
\usepackage{enumitem}
\usepackage{wrapfig,lipsum}
\usepackage{colortbl}
\usepackage{relsize}
\usepackage{ragged2e}
\usepackage{listings}
\usepackage{svg}
\definecolor{cvprblue}{rgb}{0.21,0.49,0.74}
\usepackage[pagebackref,breaklinks,colorlinks,allcolors=cvprblue]{hyperref}
\def\paperID{3306} 
\def\confName{CVPR}
\def\confYear{2026}

\title{No Calibration, No Depth, No Problem: Cross-Sensor View Synthesis with 3D Consistency}

\author{
    Cho-Ying Wu,\, Zixun Huang,\, Xinyu Huang,\, Liu Ren
    \\
    {\small Bosch Research North America \& Bosch Center for Artificial Intelligence (BCAI)}
    \\
    {\small\texttt{\{Cho-Ying.Wu, Zixun.Huang, Xinyu.Huang, Liu.Ren\}@us.bosch.edu}}\\
}

\makeatletter
\let\@oldmaketitle\@maketitle
\renewcommand{\@maketitle}{\@oldmaketitle
\vspace{-15pt}
\centering\includegraphics[trim=0mm 0mm 0mm 0mm, clip, width=0.95\linewidth]{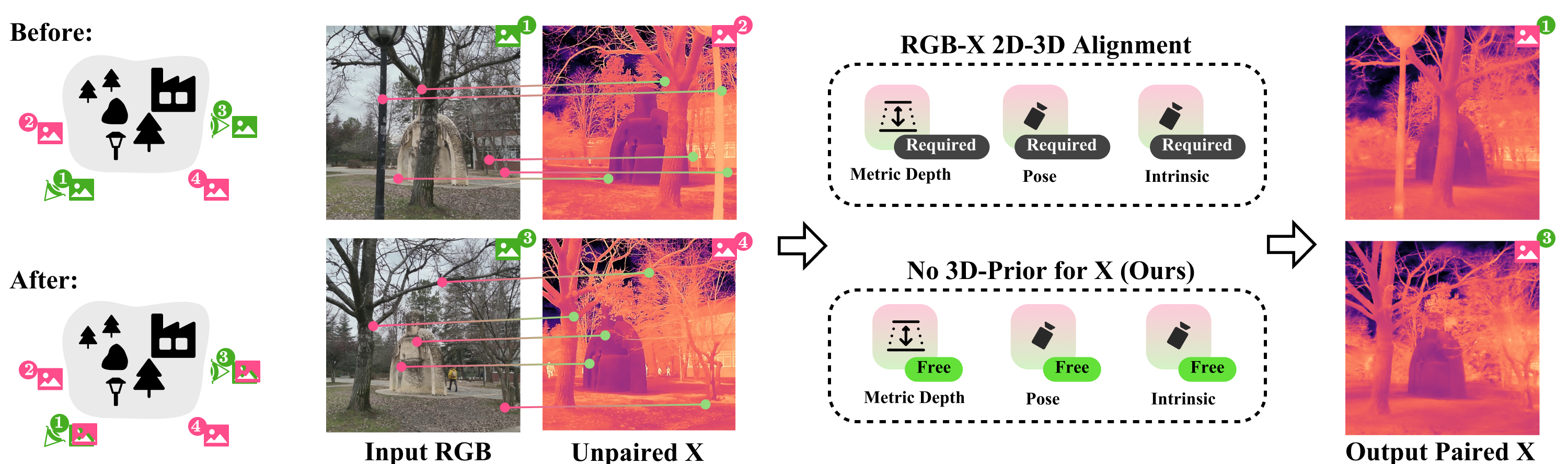}
\vspace{-5pt}
\captionof{figure}{
\textbf{Problem Setup.} Given unpaired RGB-X images from sensors, the task is to synthesize X-images that are pixel-wise aligned with the RGB views for multi-modal applications.
Traditional 3D approaches rely on complete 3D priors—including depth and the poses/intrinsics of both modalities—to align and render cross-sensor images.
In contrast, our scalable framework removes these dependencies, enabling RGB-guided X-image synthesis without the 3D priors for X to replace calibration for different types of sensors and metric depth acquisition.
}
\vspace{8pt}
\label{teaserfigure}}

\begin{document}
\maketitle

\begin{abstract}
We present the first study of cross-sensor view synthesis across different modalities. We examine a practical, fundamental, yet widely overlooked problem: getting aligned RGB-X data, where most RGB-X prior work assumes such pairs exist and focuses on modality fusion, but it empirically requires huge engineering effort in calibration. We propose a match-densify-consolidate method. First, we perform RGB-X image matching followed by guided point densification. Using the proposed confidence-aware densification and self-matching filtering, we attain better view synthesis and later consolidate them in 3D Gaussian Splatting (3DGS). Our method uses no 3D priors for X-sensor and only assumes nearly no-cost COLMAP for RGB. We aim to remove the cumbersome calibration for various RGB-X sensors and advance the popularity of cross-sensor learning by a scalable solution that breaks through the bottleneck in large-scale real-world RGB-X data collection. See \href{https://choyingw.github.io/3d-rgbx.github.io/}{project page}.  
\end{abstract}

\section{Introduction}
\label{sec:intro}
Sensors other than the RGB spectrum are useful for various applications, such as NIR (Near-Infrared) sensors for night vision in autonomous driving~\cite{choi2018kaist, li2023emergent, bustos2023systematic, munir2021sstn}, or thermal sensors for leak detection or robust autonomous systems~\cite{jadin2014gas, munir2022exploring, li2025improving, franchi2024infraparis, mirlach2025r,kim2025pixelnir}. 
Although with diverse applications, they remain relatively underexplored in research. First,
they are much less in data quantity compared with RGB, and the research on other sensors needs to rely on their paired and view-aligned RGB to provide rich information from foundation models for scene understanding~\cite{zhao2025unveiling, fan2024generalizable, paranjape2025f}. For sensors such as thermal cameras, images are natively low-texture and rely on RGB to provide features.



However, real data captured by different sensors typically suffer from multiple issues, making it challenging to build an RGB-X paired dataset, where X denotes other types of sensors in the work. 
The traditional industrial settings leverage 3D reprojection, which requires \textbf{cumbersome sensor calibration} including \textit{measuring intrinsics}, \textit{sensor synchronization}, \textit{relative pose estimation}, and \textit{metric depth}, where errors within each stage will be propagated and affect the end results, and it still cannot solve occlusion from displacement.

Structure-from-motion methods like COLMAP~\cite{schonberger2016structure} are widely used in scene reconstruction and considered much less effort in sensor calibration. Yet, it only works on RGB scenes and usually fails to work on X or RGB-X settings, particularly on less-texture sensors such as thermal cameras.

The RGB-X systems usually require additional high-precision dense depth sensors, such as RGB-Thermal-Depth system~\cite{kong2021direct}, but it further complicates sensor registration efforts and is not scalable. Some datasets simply consider fixed depth~\cite{luthermalgaussian} for re-projection without solving the rooted issue. 
Note that RGB-D systems already encode 3D information that can be re-projected to RGB's view (still requiring intrinsics and limited by depth sensor precision and density), while other sensors without such 3D priors cannot. 

\begin{figure}[t]
    \centering
    \includegraphics[width=0.95\linewidth]{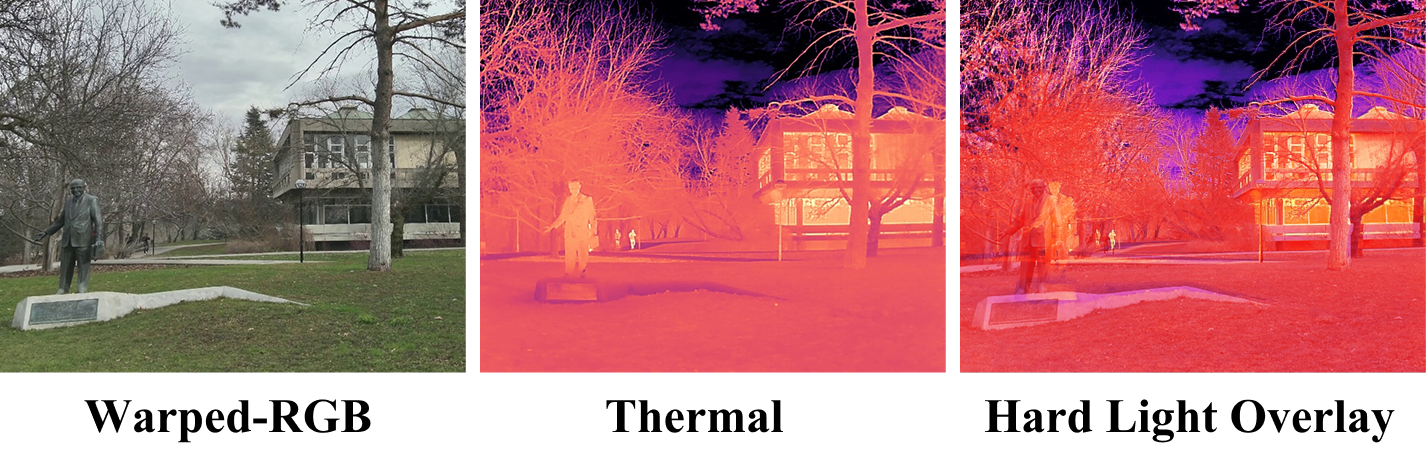}
    \vspace{-8pt}
    \caption{Homography warping assumes 3D planar structures and causes visible misalignment (statue areas) when the scene contains distinct fore-/background layers.}
    \vspace{-10pt}
    \label{fig:X2RGB_warp_error}
\end{figure}

Some recent work performs cross-modality matching of keypoints between RGB and X views ~\cite{ren2025minima, yang2025distillmatch, tuzcuouglu2024xoftr} based on RGB image matchers.  
However, they either produce sparse or incomplete correspondences or have highly noisy matches in low-confidence regions. Their goal is to estimate the camera pose or homography matrix $H \in \mathcal{R}^{3\times3}$ for warping, where a target view pixel $p'$ and source view pixel $p$ (both in homogeneous coordinates) are related by $p'=H^{-1}p$.
However, homography warping assumes \textit{3D planar structures}, indicating it can only distort or shear the view as a plane but cannot create 3D effects with disparity. The warping works when the scene lies around a depth layer, but will show distinct misalignment in counterexamples (See Fig.~\ref{fig:X2RGB_warp_error}).

This work is \textbf{the first scalable framework for cross-sensor view synthesis} to acquire paired and aligned RGB-X data without cumbersome calibration or metric depth, to facilitate the research on sensors \textbf{without 3D priors and beyond RGB cameras} (Fig.~\ref{teaserfigure}), such as \textit{Thermal Cameras}, \textit{NIR Cameras}, and \textit{Synthetic-Aperture Radar (SAR)}. 

First, we adopt cross-modal image matchers to match keypoints between RGB and X views and accumulate X's keypoints across frames onto the current RGB view to form an aligned sparse or semi-dense X-map. 
Then, both the RGB and X-map are input to our trained RGB-guided X-densifier, which reconstructs the dense X-map.
With a handful of keypoints aggregated in X-map and RGB images as guidance, the model generates a dense X-image that satisfies the input X-map and infills the void areas using RGB and surrounding points as cues.
However, we find that densification may lead to irregular or noisy structures affected by noisy matches in low-confidence areas. 
A higher confidence threshold yields a sparser X-map that must rely more on cross-domain guidance from RGB, while a lower threshold introduces noisy X-points and distorts structures.
We propose \textbf{Confidence-Aware Densification and Fusion (CADF)} that densifies and fuses multi-level threshold X-maps into one X-image.

Then, we perform patch rejection by our \textbf{self-matching} mechanism---We look into the patch features from the image matcher with the prior that each patch should be matched to the same patch for RGB-X pairs. We calculate a similarity matrix between each patch and reject low-similarity patches. After that, we perform a finer-stage densification. To further enhance multi-view consistency, we adopt RGB-X 3D Gaussian Splatting (3DGS) on RGB's views to consolidate in the 3D space for multi-view consistency. The work only considers running COLMAP on RGB for 3DGS training, which is standard and considered no-cost. In the experiments, we further show that even without using 3DGS, we still attain better performance than others.

\textbf{Contributions}. Our contributions are as follows.
\begin{itemize}
\item The first scalable framework for cross-sensor view synthesis on RGB-X to get view-aligned pairs without calibration or depth---studying a fundamental but widely overlooked problem in prior cross-modal/sensor learning.
\item A \textit{match-densify-consolidate} framework. It begins with matched sparse keypoints and presents a densification approach with our CADF module that integrates matching confidence into the densifier for better synthesis, followed by subsequent filtering, re-densification, and 3DGS to build up 3D consistency.
\item Extensive evaluation demonstrating our method attains the state of the art on the methods without 3D priors, even still outperforming them without leveraging 3DGS.
\end{itemize}

\section{Related Work}
\label{sec:intro} 

\subsection{RGB-X Task and Data Curation}
Imaging sensors beyond RGB like NIR or thermal cameras are useful in industrial applications, especially in autonomous driving for adverse weather or night vision to ensure safety~\cite{palladin2024samfusion,bijelic2020seeing, shaik2024idd,shin2023deep, kim2024exploiting,ji2023multispectral}. 
Some works study sensor fusion strategy for segmentation, detection, or tracking~\cite{jin2022darkvisionnet,sun2019rtfnet,liu2023multi,zhang2023cmx,tang2024divide,zhou2021gmnet, wan2025sigma,liu2024infrared,zheng2024learning,zhang2021abmdrnet,deng2021feanet,xu2021multimodal,deevi2024rgb,frigo2022doodlenet,wu2021scene,li2020challenge,hong2024onetracker,hou2024sdstrack,hui2023bridging,zhang2023efficient} and applications in robotics~\cite{liang2023explicit,feng2023cekd,madan2024rabbit,zhang2022visible,aditya2024thermal,jiang2022thermal}

Most works focus on sensor fusion and design fusion modules. However, they all assume \textbf{pixel-wise aligned RGB-X inputs already}, which require substantial engineering effort in sensor calibration, synchronization, and precise metric depth from sensors like Lidar or ToF to help alignment. Some works adopt image translation from RGB to generate X-images, such as RGB-to-thermal, to get pseudo-paired data~\cite{li2025pseudo,xiao2025thermalgen,abbott2020unsupervised,berg2018generating}. They do not use real information in X, and the generation needs to deal with inherent ambiguity. E.g., a cup of water can be flexible in its exact temperature (cool or cold) and hard to judge by its appearance.

In contrast to prior works, we focus on this fundamental problem in data and propose a scalable framework to reduce the engineering effort to obtain paired RGB-X real data.

\subsection{Cross-Modal Image Matching}

Image matching is a fundamental problem. Early methods adopt handcrafted descriptors~\cite{brown2005multi,tola2008fast} and find a transformation with RANSAC~\cite{fischler1981random}. During the deep learning era some works adopt CNN~\cite{wang2021p2,tyszkiewicz2020disk,ono2018lf,dusmanu2019d2, edstedt2023dkm}, graph neural network~\cite{sarlin2020superglue,zhang2019deep,shi2022clustergnn}, transformers~\cite{sun2021loftr,ni2024eto,wang2022matchformer,chen2022aspanformer}, lightweight attention~\cite{lindenberger2023lightglue,potje2024xfeat}, or benefit from foundation models~\cite{Edstedt_2024_CVPR,Jiang_2024_CVPR}. 
They focus on matching in the RGB domain from different views, and performance severely degrades for cross-modal matching. 
Further, image matching aims for relative pose estimation and still needs 3D priors such as intrinsics and precise metric depth for re-projection. Without 3D priors, they still use an estimated homography matrix to warp to the target view, which is confined by its underlying assumption (Fig.~\ref{fig:X2RGB_warp_error}).

Recent feed-forward reconstruction methods, like DUSt3R~\cite{wang2024dust3r}, VGGT~\cite{wang2025vggt}, or MapAnything~\cite{keetha2025mapanything}, also implicitly predicts intrinsics, depth, and 3D pointmap and reconstructs the scene.
However, they cannot perform cross-modal matching, and reconstruction would fail due to feature dissimilarity.  

Few works have explored cross-modal matching. ReDFeat~\cite{deng2022redfeat} uses an adapter and detector to get features and compute detection scores for RGB-thermal pairs.
XoFTR~\cite{tuzcuouglu2024xoftr} uses transformers for coarse-to-fine RGB-thermal matching and achieves stronger performance.
MINIMA~\cite{ren2025minima} presents a cross-modal image generation engine (e.g., RGB-to-thermal) to finetune image matchers, such as LightGLUE~\cite{lindenberger2023lightglue} as its featured model, thereby enhancing feature robustness across modalities. However, they still estimate poses or homography matrices and remain confined by the assumptions behind.

\begin{figure*}[t]
  \centering
  \includegraphics[width=0.95\textwidth]{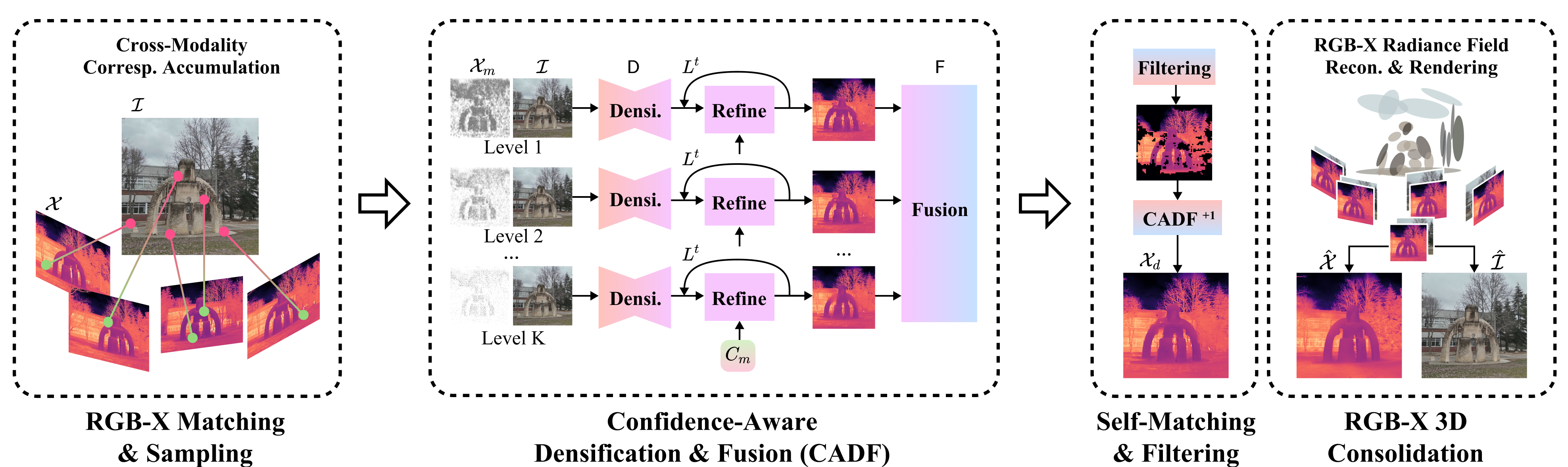}
\caption{\textbf{Method Overview.} 
Our approach consists of three stages. 
In the first stage, we perform \textit{cross-modality feature matching} to establish correspondences between RGB and X-images. 
The matched points are sampled and accumulated onto RGB views to produce semi-dense X-images $\mathcal{X}_m$ along with multi-level confidence maps $C_{m}$. 
In the second stage, we conduct RGB-guided densification to get dense X-images from RGB with semi-dense X as cues. Our Confidence-Aware Densification and Fusion module integrates confidence maps from the image matching stage to guide the densification to concentrate on higher-confidence points and get robust $\mathcal{X}_d$
In the final stage, our proposed \textit{self-matching mechanism} further filters inconsistent patches, and the results are fed back into the densification stage for refinement. 
To further improve multi-view consistency, we train RGB-X 3DGS using COLMAP-calibrated RGB views to consolidate both modalities into a unified 3D RGB-X radiance field, which improves multi-view consistency and further enables cross-sensor view synthesis.  
}
\label{fig:method}
\vspace{-9pt}
\end{figure*}

\section{Methods}


We enable RGB-X cross-sensor view synthesis without requiring 3D priors for X, such as scene metric depth and calibration for cross-sensor relative pose and intrinsics (Fig.~\ref{teaserfigure}).

Detailed method is shown in Fig.~\ref{fig:method}. We begin with the smallest components--- pixel-level keypoints from a cross-modal image matcher--- and perform RGB-X densification to generate X-images aligned with RGB. 
Our Confidence-Aware Densification and Fusion (CADF) module integrates image-matching confidence into the densification stage to achieve better synthesis. 
Then, we propose self-matching for filtering, where we use the matcher as an evaluator to filter out erroneous patches, followed by finer-stage re-densification.
Lastly, we employ RGB-X 3DGS using RGB camera poses along with densified and aligned X-view, and consolidate both sensors into an unified 3D space. Notably, the pipeline only requires cheap and widely-used COLMAP on RGB.
We show that even without 3DGS and COLMAP, our densified maps still outperform baselines in experiments. 

\subsection{RGB-X Matching}

\textbf{Matching}. Given an RGB image $\mathcal{I}$ and X-modality image $\mathcal{X}$, image matching finds a matching set 
that maps coordinates $p^\mathcal{I}$ in $\mathcal{I}$ to $p^\mathcal{X}$ in $\mathcal{X}$ with a confidence score $c$. 
Most matchers can afford sparse or semi-dense matching in practice, where memory consumption directly scales with the number of candidates. 
The matching results are filtered by a threshold $\delta$, where a match is kept when $c >= \delta$.

Cross-modal matchers typically yield far fewer correspondences than RGB-based ones.
This is partly due to the inherent challenge of aligning features across modalities.
Further, modalities like thermal or SAR often contain large homogeneous regions (e.g., uniform-temperature areas lacking surface texture), making it difficult to extract feature points for reliable matching. 

We begin from an image matcher and stack $N$ frames' X-keypoints into corresponding RGB's coordinates to get ${\mathcal{X}_m}$.
\begin{equation}
\label{aggregating_N_frames}
{\mathcal{X}_m}[p] 
= 
\frac{
    \sum_{n} \mathbf{1}[p = p_n^{\mathcal{I}}] \, \mathcal{X}[p_n^{\mathcal{X}}]
}{
    \sum_{n} \mathbf{1}[p = p_n^{\mathcal{I}}]
},
\end{equation}
where $n \in N$, and $\mathbf{1}[\cdot]$ is the indicator function that equals 1 if the condition holds and 0 otherwise. If $p \neq p^\mathcal{I}_n, \forall n \in N$, ${\mathcal{X}_m}[p] = -1$, meaning void area.

\noindent\textbf{Area Sampling}.
To ameliorate matching in texture-less or complex areas, such as sky, ground, walls, or grass, we adopt GroundedSAM~\cite{ren2024grounded} to segment these areas on RGB images. 
Then, we uniformly sample points from the corresponding warped X-images (via homography) within the mask and still void areas. 
To prevent warping errors from heavily influencing subsequent densification, and to allow erroneous matches to be filtered out later, we only sample 5\% points of such areas.
{\small
\begin{equation}
\label{area_sampling}
{\mathcal{X}_m}[p] = {\mathcal{X}_W}[p],\;\;
p \sim \mathrm{U}\big(\{ p \mid \mathcal{M}(p) = \text{1}  \wedge {\mathcal{X}_m}[p] = \text{-1}\}\big),
\end{equation}
}
where $\mathrm{U}$ is uniform sampling, ${\mathcal{X}_W}$ denotes the warped X-image to RGB view, using a homography matrix computed from RGB-X correspondences, and $\mathcal{M}$ is the area mask.

\subsection{Confidence-Aware Densification and Fusion}
\textbf{Densification}. The obtained ${\mathcal{X}_m}$ is sparse/ semi-dense, and then we densify it by a network $\textsf{D}$ to get a dense map. 
$\textsf{D}$ is pretrained on diverse RGB–X modalities using paired data, where the model takes an RGB image along with a downsampled, sparsified X-map as inputs. The model is trained to densify X into a dense map.
The architecture contains a backbone of recurrent units~\cite{zuo2025omni, zuo2024ogni} and dynamic spatial propagation (DySPN) layers~\cite{lin2022dynamic} that refine the output with spatial attention from known points in a recurrent fashion.
For inference, $\mathcal{I}$ and ${\mathcal{X}_m}$ are inputs. 
Using a handful of points as cues in ${\mathcal{X}_m}$, $\textsf{D}$ generates a better dense map that conforms to known conditions ${\mathcal{X}_m}$ and fills in unknown areas from $\mathcal{I}$ and neighboring points.

\noindent\textbf{Confidence-Aware Fusion}. 
Naively densifying ${\mathcal{X}_m}$ would lead to highly irregular structures due to uncertainty in image matching.
To overcome the issue, we design Confidence-Aware Densification and Fusion (CADF) module that integrates the matching and densification stages by using the image-matching confidence $c$ to improve the RGB-X densification.

First, we integrate $c$ into DySPN refinement. The original DySPN formulates the recurrent iteration as follows: at timestep $t$, the refined result $L^{t+1}$ combines current update $L^t$, weighted by affinity score $w$, and known points $X_m$, weighted by backbone-predicted certainty map $C_s$
\begin{equation}
\label{spn_conf}
L^{t+1} = (1-C_{s}) \sum_r \sum_{(a,b)} w_{r,a,b} * L^t_{a,b} + C_{s}  {\mathcal{X}_m},
\end{equation},
where $r$ denotes filters and $(a,b)$ denotes the coordinates.


We propose to interleave the recurrent steps with the confidence map $C_{m}$ from image matching 
that aggregates point confidence $c$ into a map.
$C_m$ weighs down the contribution for probably incorrect original keypoints and concentrates the iterative refinement at the higher-confidence points.
{\small
\begin{equation}
\label{blend_conf}
L^{t+1} = (1-C_{s}C_m) \sum_r \sum_{(a,b)} w_{r,a,b} * L^t_{a,b} + C_{s}C_m  {\mathcal{X}_m}.
\end{equation}
}

Next, we find that a higher $\delta$ keeps more robust points but leads to a sparser map that purely relies on RGB as cues for densification. A lower $\delta$ accepts more but noisy points as known conditions, which leads to irregular structures (See Supplementary for visuals). We perform $K$-level thresholding and fusion.
Suppose we have $K$ thresholds $\delta_k$ for confidence $c$, thresholded keypoints ${\mathcal{X}_{m,k}}$, and their densified X-images ${\mathcal{\hat{X}}_{d,k}}, \forall k \in [1,K]$, we propose a fusion block $\textsf{F}$ at the end of densification to fuse them. 
$\textsf{F}$ is first pretrained on single-image enhancement to suppress noise, deblur, and sharpen edges. 
${\mathcal{\hat{X}}_{d,k}}$ are input to $\textsf{F}$, and we mean pool over $k$ to output ${\mathcal{X}_{d}}$, as the result for the CADF stage.
We train $\textsf{F}$ with self-supervised losses, including cosine similarity loss (Eq.~\ref{loss_cosine_sim}) using SigLIP2 image encoder~\cite{tschannen2025siglip} along with RGB-X self-matching loss (Eq.~\ref{loss_trace})---both using RGB images to guide the enhancement and fusion for X-images.
\begin{equation}
\label{loss_cosine_sim}
\mathcal{L}_{\text{cos}}(\mathcal{I}, {\mathcal{X}_{d}})
= 1 - \frac{f_{\text{SigLIP}}(\mathcal{I})^{\top} f_{\text{SigLIP}}({\mathcal{X}_{d}})}
{\|f_{\text{SigLIP}}(\mathcal{I})\|_2 \, \|f_{\text{SigLIP}}({\mathcal{X}_{d}})\|_2},
\end{equation}
where both RGB and X are capturing the same scene that could be matched with the same descriptions, and thus the cosine similarity loss helps maximize feature map similarity.

\begin{figure*}[h]
    \centering
    \includegraphics[width=0.92\linewidth]{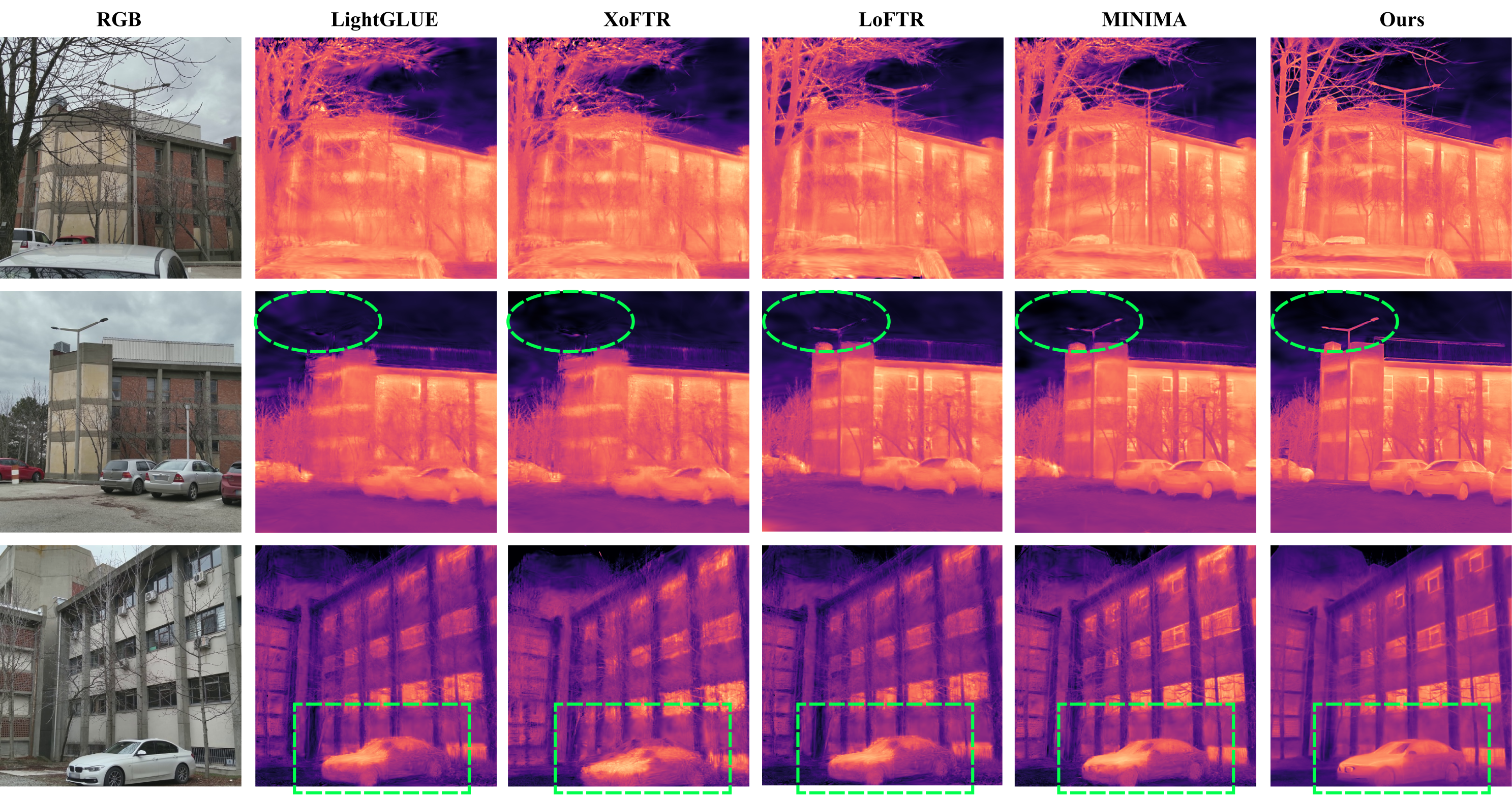}
    \vspace{-4pt}
    \caption{\textbf{Visual Results on METU-VisTIR-Cloudy.} Our results attain much clearer, sharper, and smoother surface for rendering.}
    \vspace{-5pt}
    \label{fig:METU_results}
\end{figure*}

\subsection{Self-Matching Filtering and 3D Consolidation}
\noindent\textbf{Self-Matching}. We propose a self-matching mechanism used both in fusion and the following filtering stage. Different from image matching--- given two images, find a matching set that maps coordinates from one to another--- now we have roughly aligned RGB-X pairs with the prior that coordinates in one image should map to the same locations in another. We can use the image matcher as a judge to evaluate whether the densified X-image is valid for matching to the same coordinates.

For training $\textsf{F}$, we look into the patch features from the transformer-based matcher and compute patch-level similarity matrix $A$, which is computed by scaled dot product of the RGB-X features.
\begin{equation}
\label{loss_trace}
A = \frac{F_{\mathcal{I}} F_{\mathcal{X}}^\top}{\tau}
,
\end{equation}
where $\tau$ is a scale factor, and $F_{\mathcal{I}}$ and $F_{\mathcal{X}}$ are features for RGB and X in the coarse matching layer of the matcher~\cite{tuzcuouglu2024xoftr}.

The ideal similarity matrix should be diagonal. We compute the loss that maximizes the diagonal sum and minimizes the off-diagonal parts by
\begin{equation}
\label{loss_trace}
\mathcal{L}_{\text{sim}}(A)
= 
- \frac{\operatorname{Tr}(A)}{\|A\|_F}
+ \lambda \frac{\|A \odot (\hat{\mathbf{1}} - I)\|_1}{\|A\|_F }
,
\end{equation}
where $\|.\|_F$ is the Frobenius norm, $\operatorname{Tr}(\cdot)$ is trace for a matrix, $I$ is the identity matrix, $\lambda$ is a weight, and $\hat{\mathbf{1}}$ is the one-matrix.

\noindent\textbf{Filtering}. We further use $A$ to reject wrongly densified patches in ${\mathcal{X}_{d}}$. 
Looking into the diagonal of $A$, we compute a measure of concentration $q = \frac{Q_{50}(\mathbf{A})}{Q_{99}(\mathbf{A})}$, with $Q(\cdot)$ as the quantile function. A higher $q$ indicates stronger self-matching results and thus fewer patches need to be rejected, while a lower $q$ means the opposite. We take (1-$q$)-quantile of $A$'s diagonal as the threshold and filter out patches of smaller scores.

\noindent\textbf{Fine-Stage Densification}.
We then perform the fine-stage densification based on the filtered X-images. We follow the previous stage to perform densification without $K$-levels, and in Eq.~\ref{blend_conf} we take the normalized similarity score from $A$ as $C_m$.

\noindent\textbf{RGB-X 3DGS}.
To further make per-frame densification 3D-consistent, we consolidate the densified X-images by 3DGS~\cite{kerbl20233d,yu2024mip,huang3dgeer}. We train RGB-X 3DGS, where we further add X-channels for each Gaussian. Unlike~\cite{katragadda2025online, peng3d}, where they render multi-modality (e.g., color and language embedding) from separate channels and disentangle the GS parameters, here we just keep one set of parameters. Because we find raw sensor imaging is often noisy or low-resolution, which cannot match the higher quality of RGB, and the latter can better position each 3D Gaussian precisely. 

\section{Experiments}
\begin{table}[t]
\centering
\small
\setlength{\tabcolsep}{1.8pt}
\caption{\textbf{Results on METU-VisTIR-Cloudy}. Results are the mean of all six sequences. We compare with warping by different image matchers, all trained and rendered by 3DGS.}
\vspace{-3pt}
\begin{tabular}{|l|c|c|c|c|c|c|c|}
\hline
\rowcolor{gray!10}
\textbf{Method} & \textbf{Icos}	
$\uparrow$ & \textbf{p30}$\uparrow$ & \textbf{p50}$\uparrow$ & \textbf{p70}$\uparrow$ & \textbf{p90}$\uparrow$ & \textbf{ITM}$\uparrow$ & \textbf{ITcos}$\uparrow$ \\
\hline
XoFTR~\cite{tuzcuouglu2024xoftr} & 0.62 & 25.13 & 27.49 & 29.31 & 31.48 & 0.69 & 0.39 \\
LightGlue~\cite{lindenberger2023lightglue} & 0.61 & 25.89 & 28.28 & 30.14 & 32.35 & 0.91 & 0.40 \\
LoFTR~\cite{sun2021loftr} & 0.66 & 29.38 & 32.07 & 33.95 & 36.04 & 0.89 & \textbf{0.45} \\
MINIMA$_{lg}$~\cite{ren2025minima} & 0.67 & 29.93 & 32.78 & 34.72 & 36.99 & 0.88 & 0.44 \\
Ours & \textbf{0.69} & \textbf{31.18} & \textbf{34.39} & \textbf{36.43} & \textbf{38.72} & \textbf{0.92} & \textbf{0.45} \\
\hline
\end{tabular}
\vspace{-3pt}
\label{tab:cloudy_rgbt_mean}
\end{table}

\begin{figure*}[bt!]
    \centering
    \includegraphics[width=0.92\linewidth]{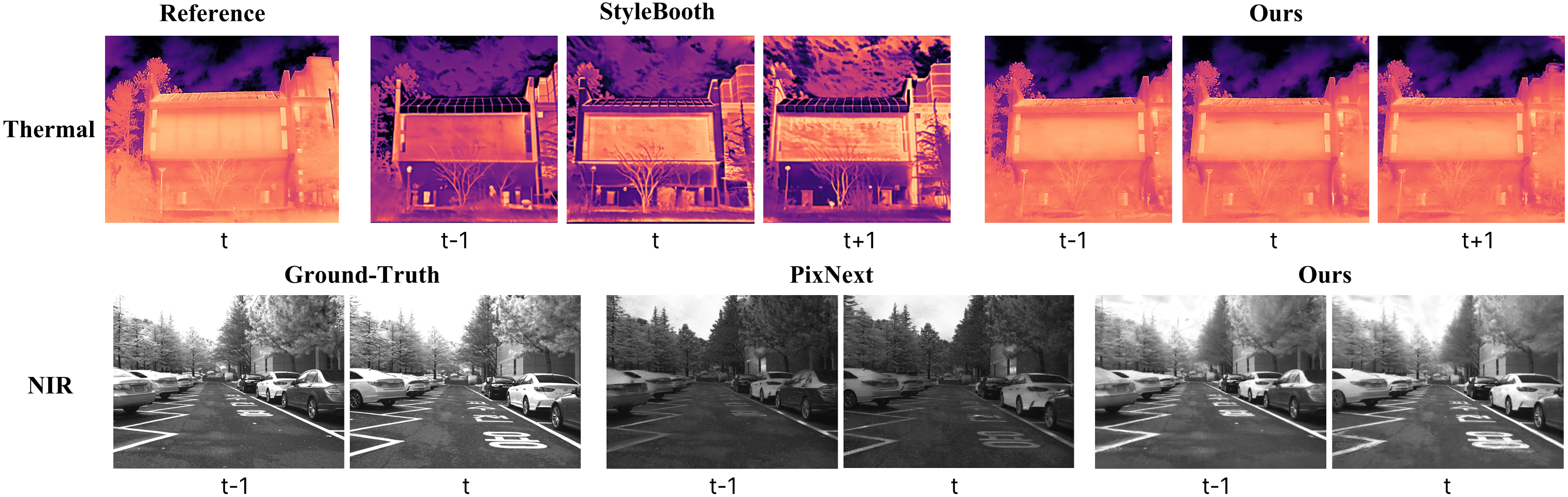}
    \vspace{-10pt}
    \caption{\textbf{Comparison on Temporal Consistency for Image Generation.} StyleBooth~\cite{Han_2025_ICCV} generation for thermal images cannot guarantee temporal consistency due to inherent ambiguity, while ours densification creates more consistent multi-views. NIR is closer to the visual spectrum and thus easier to ensure consistency, but the specialized method PixNext~\cite{jin2025pix2next} still cannot ensure the correct intensity. Compared with our strategy, image translation from the original domain still suffers from inaccurate transformations, whereas our match-densify-consolidate approach uses information from the target domain as anchors for densification and achieves better results. 
    }
    \vspace{-5pt}
    \label{fig:temporal_consistency_rgbnir}
\end{figure*}




\noindent\textbf{X-Modalities and Datasets}. 

\noindent1. \textbf{RGB-Thermal}. The popular METU-VisTIR-Cloudy Dataset~\cite{tuzcuouglu2024xoftr}, which includes six sequences captured by \textit{raw and unpaired} RGB and thermal sensors, is adopted for testing. We also use four scenes from RGBT-Scenes~\cite{luthermalgaussian} containing paired RGB-thermal data for evaluation. We apply random rotation, shift, perspective warping, and radial distortion to create unpaired RGB-thermal data. To train $\textsf{D}$ for this modality, we use synthetic RGB-thermal pairs created by MINIMA~\cite{ren2025minima}, which uses the MegaDepth Dataset~\cite{li2018megadepth} to create a large volume of paired data.

\noindent2. \textbf{RGB-NIR}. We adopt five sequences from RGB-NIR-Stereo~\cite{kim2025pixelnir}, including one shorter and four extracted from two much longer sequences. We follow the same style as RGBT-Scenes to create unpaired data. To train $\textsf{D}$, we adopt Deep-NIR~\cite{sa2022deepnir} with synthetic paired RGB-NIR data.

\noindent3. \textbf{RGB-SAR}. We adopt three sub-Gigapixel satellite RGB-SAR pairs from DDHR-HK~\cite{ren2022dual} and cut them into 512x512 patches for evaluation. We do not perform RGB-X 3DGS only on this modality since satellite images inherently cannot capture multi-view 3D.
Several RGB-SAR datasets~\cite{cardona2024dataset,xiang2020automatic,zhao2025remotesensingdataset,li2022mcanet} are used to train $\textsf{D}$.

\noindent\textbf{Implementation Details.} We adopt the transformer-based XoFTR~\cite{tuzcuouglu2024xoftr} as the image matcher, which is trained on RGB-thermal but also presents strong ability for across different spectrum~\cite{ren2025minima}. 
We choose $N=7$ (-3 to +3 frame\footnote{The work reasonably assumes rough alignment. A standard RGB-X data capture can trigger RGB-X shutters around the same time with similar scene coverage, but hard to be exact or aligned, as this is affected by on-the-fly system loading, physical displacement of sensors, or different intrinsics.}), $K=3$ levels, $\delta=0.15, 0.3, 0.5$, $\lambda=0.1$, $\tau=0.1$, and
$c=0.3$ for those area-sampled points.
Fusion module $\textsf{F}$ is an image-enhancement network~\cite{Wang_2022_CVPR} and pretrained on DIV2K~\cite{Agustsson_2017_CVPR_Workshops} for image denoising, deblurring, and super-resolution.
See Supplementary for details of training $\textsf{D}$ and $\textsf{F}$.

 

\begin{figure}[bt!]
    \centering
    \includegraphics[width=0.96\linewidth]{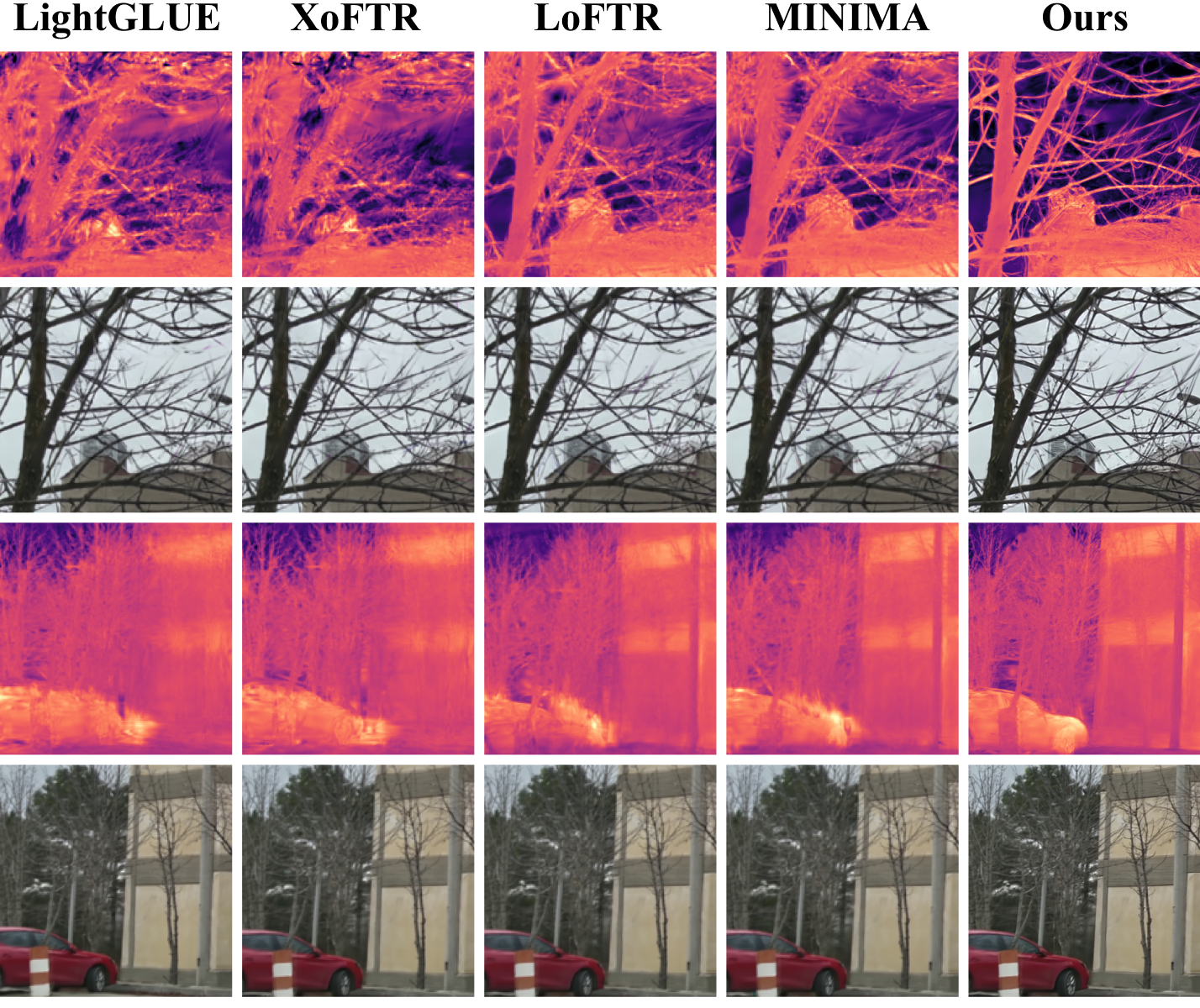}
    \vspace{-8pt}
    \caption{With the aid of sharper and clearer thermal images, the RGB view synthesis quality is also slightly enhanced.}
    \vspace{-5pt}
    \label{fig:X2RGB_side_effect}
\end{figure}

\begin{table}[tb!]
\centering
\small
\setlength{\tabcolsep}{2pt}
\caption{\textbf{Comparison on Temporal Consistency}. Compared with pure image generation, our match-densify-consolidate strategy obtains better multiview consistency from \textbf{lower} MEt3R score.}
\vspace{-3pt}
\begin{tabular}{|l|c|c|c|c|c|c|c|}
\hline
\rowcolor{gray!10}
\textbf{MEt3R$\downarrow$} & \textbf{Sc1} & \textbf{Sc2} & \textbf{Sc3} & \textbf{Sc4} & \textbf{Sc5} & \textbf{Sc6} & Mean \\
\hline
StyleBooth~\cite{Han_2025_ICCV} & 0.202 & 0.331 & 0.357 & 0.285 & 0.289 & 0.319 & 0.297 \\
Ours & \textbf{0.141} & \textbf{0.123} & \textbf{0.143} & \textbf{0.258} & \textbf{0.158} & \textbf{0.202} & \textbf{0.171} \\
\hline
\end{tabular}
\vspace{-5pt}
\label{tab:metu_met3r}
\end{table}

\begin{figure*}[bt!]
    \centering
    \includegraphics[width=0.96\linewidth]{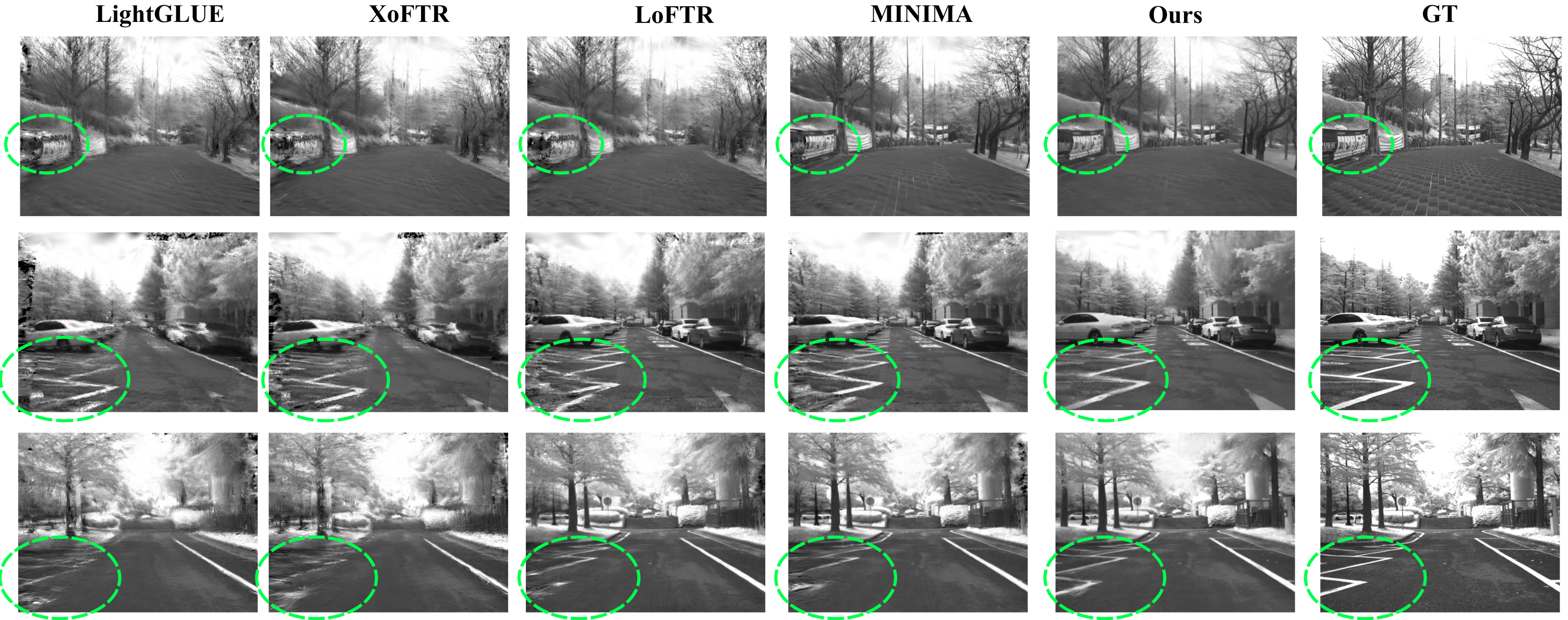}
    \vspace{-7pt}
    \caption{\textbf{Visual Results on RGB-NIR-Stereo.} Our view synthesis showcases better structures closer to the groundtruth (GT).}
    \vspace{-5pt}
    \label{fig:NIR_results}
\end{figure*}

\noindent\textbf{Metrics.}
For datasets with only unpaired sequences without groundtruth, like METU-VisTIR-Cloudy, we calculate (1) image cosine similarity (Icos) for RGB-X image features from SigLIP-2 image encoder~\cite{tschannen2025siglip}, (2) 30/50/70/90-percentile (p30-p90) on the diagonal of the similarity matrix of XoFTR for RGB-thermal, (3) image captioning and matching score--- We use popular MLLM BLIP-2~\cite{li2023blip} to first caption the RGB image and append ", in a thermal image." to the text. The caption and thermal images are used to calculate image-text matching score (ITM) and image-text cosine similarity (ITcos). Specifically, the match/mismatch score comes from
a classification head that predicts two logits $z_{\text{match}}$ and $z_{\text{mismatch}}$ from its linear alignment. Then $\mathrm{prob}_{\text{mismatch}} = \frac{e^{z_{\text{mismatch}}}}{e^{z_{\text{match}}} + e^{z_{\text{mismatch}}}}.$

For datasets with groundtruth, we calculate standard PSNR, SSIM, LPIPS as the fidelity scores, or RMSE/ MAE if the maps carry physical units such as $^\circ$C. To evaluate temporal consistency, we use MEt3R~\cite{asim2025met3r}, which \textit{lower scores} indicate \textit{better multi-view consistency}.

\noindent\textbf{Baselines.} We include homography warping from different image matchers, including XoFTR~\cite{tuzcuouglu2024xoftr}, LightGLUE~\cite{lindenberger2023lightglue} LoFTR~\cite{sun2021loftr}, and MINIMA$_{lg}$~\cite{ren2025minima} as the featured matcher, followed by 3DGS training using their warped views. We also include cross-modal image generation, where models are trained with specific modalities, such as StyleBooth~\cite{Han_2025_ICCV} used in ~\cite{ren2025minima} for RGB-thermal and~\cite{jin2025pix2next} for RGB-NIR. 

The work compares against the above methods without 3D priors for X, which aligns with our scope. 
We also consider 3D reprojection with estimated metric depth (from DepthAnythingV2~\cite{yang2024depth}), estimated camera pose (from essential matrix using matched keypoints), and groundtruth intrinsics for X. However, the pipeline cannot establish a reasonable baseline because metric depth and cross-modal camera pose estimation are far from robust in the wild~\cite{wu2023inspacetype, Wu_2022_CVPR, wu2024boosting, guo2025depth}.

\subsection{RGB-Thermal}

\textbf{METU-VisTIR-Cloudy}. The dataset contains unpaired RGB and thermal sequences without groundtruth, and we adopt several metrics calculated against color images instead. See \textbf{Metrics}. 
From Tab.~\ref{tab:cloudy_rgbt_mean}, our method consistently performs the best across all metrics, showing superior view-synthesis quality and faithful alignment with associated images. Especially for p30-p90, the higher similarity scores validate that our X-images make it easier for a pretrained image matcher to build correspondences. Fig.~\ref{fig:METU_results} shows a visual comparison, where our results show the most clear object structures of the thermal images. Further, we find that we also attain sharper RGB view synthesis (Fig.~\ref{fig:X2RGB_side_effect}) than others due to better thermal images used in 3D consolidation.

We then compare with image generation from RGB views. We follow MINIMA~\cite{ren2025minima}, which uses StyleBooth~\cite{Han_2025_ICCV} to craft paired RGB-thermal. Inherent ambiguity between appearance and true temperature makes it hard to estimate the exact temperature only from the appearance.
This makes it unstable to generate consistent thermal images, while ours acquires true values by image matching and anchors those points in densification to synthesize views conforming to the true values.
Tab.~\ref{tab:metu_met3r} shows that StyleBooth presents worse multiview consistency, where higher MEt3R~\cite{asim2025met3r} indicates worse consistency. Fig.~\ref{fig:temporal_consistency_rgbnir} shows the differences in the upper.



\noindent\textbf{RGBT-Scenes}. 
We follow the official split and report train and novel view synthesis results. Since the dataset contains paired RGB-thermal views with real temperatures, we compute RMSE and MAE in $^\circ\mathrm{C}$ against groundtruth. Tab.~\ref{tab:rgbt_scenes} displays the results, where our method shows lower errors in nearly all sequences. 

\begin{table}[t]
\centering
\footnotesize
\setlength{\tabcolsep}{1.0pt}
\caption{\textbf{Comparison on RGBT-Scenes.} We show results for train and novel view synthesis.
Each scene reports RMSE$\downarrow$ and MAE$\downarrow$ in $^\circ$C.
Novel views from the official split contain simpler views with lower errors than train views.}
\begin{tabular}{|l|cc|cc|cc|cc|cc|}
\hline
\rowcolor{gray!10}
\textbf{Train view} & \multicolumn{2}{c|}{\textbf{Building}}  & \multicolumn{2}{c|}{\textbf{Landscape}} & \multicolumn{2}{c|}{\textbf{Parterre}} & \multicolumn{2}{c|}{\textbf{RoadBlock}} & \multicolumn{2}{c|}{\textbf{Mean}} \\
\cline{2-11}
\rowcolor{gray!10}
 & \textbf{{\smaller[2]{RMSE}}} & \textbf{{\smaller[2]{MAE}}} 
 & \textbf{{\smaller[2]{RMSE}}} & \textbf{{\smaller[2]{MAE}}} 
 & \textbf{{\smaller[2]{RMSE}}} & \textbf{{\smaller[2]{MAE}}} 
 & \textbf{{\smaller[2]{RMSE}}} & \textbf{{\smaller[2]{MAE}}} 
 & \textbf{{\smaller[2]{RMSE}}} & \textbf{{\smaller[2]{MAE}}} \\
\hline
XoFTR~\cite{tuzcuouglu2024xoftr} & 2.50 & 1.81 & 1.08 & 0.79 & 2.25 & 1.57 & 1.21 & 0.93 & 1.76 & 1.28 \\
LightGLUE~\cite{lindenberger2023lightglue} & 2.48 & 1.79 & 1.10 & 0.81 & 2.28 & 1.57 & 1.26 & 0.97 & 1.78 & 1.29 \\
LoFTR~\cite{sun2021loftr} & 2.44 & 1.75 & 1.06 & 0.77 & 2.20 & 1.52 & 1.36 & 0.99 & 1.77 & 1.26 \\
MINIMA~\cite{ren2025minima} & 2.51 & 1.83 & 1.08 & 0.80 & 2.35 & 1.66 & 1.36 & 1.04 & 1.83 & 1.33 \\
\textbf{Ours} & \textbf{2.43} & \textbf{1.72} & \textbf{1.02} & \textbf{0.71} & \textbf{2.18} & \textbf{1.49} & \textbf{1.18} & \textbf{0.91} & \textbf{1.70} & \textbf{1.21} \\
\hline
\end{tabular}

\begin{tabular}{|l|cc|cc|cc|cc|cc|}
\hline
\rowcolor{gray!10}
\textbf{Novel View} & \multicolumn{2}{c|}{\textbf{Building}}  & \multicolumn{2}{c|}{\textbf{Landscape}} & \multicolumn{2}{c|}{\textbf{Parterre}} & \multicolumn{2}{c|}{\textbf{RoadBlock}} & \multicolumn{2}{c|}{\textbf{Mean}} \\
\cline{2-11}
\rowcolor{gray!10}
 & \textbf{{\smaller[2]{RMSE}}} & \textbf{{\smaller[2]{MAE}}} 
 & \textbf{{\smaller[2]{RMSE}}} & \textbf{{\smaller[2]{MAE}}} 
 & \textbf{{\smaller[2]{RMSE}}} & \textbf{{\smaller[2]{MAE}}} 
 & \textbf{{\smaller[2]{RMSE}}} & \textbf{{\smaller[2]{MAE}}} 
 & \textbf{{\smaller[2]{RMSE}}} & \textbf{{\smaller[2]{MAE}}} \\
\hline
XoFTR~\cite{tuzcuouglu2024xoftr} & 2.70 & 1.92 & 1.02 & 0.70 & 1.34 & 0.84 & 0.47 & 0.40 & 1.38 & 0.97 \\
LightGLUE~\cite{lindenberger2023lightglue} & 2.78 & 2.00 & 1.01 & 0.73 & 1.21 & 0.71 & 0.85 & 0.68 & 1.46 & 1.03 \\
LoFTR~\cite{sun2021loftr} & 2.89 & 2.11 & 1.02 & 0.71 & 1.47 & 1.01 & 0.69 & 0.49 & 1.52 & 1.08 \\
MINIMA~\cite{ren2025minima} & 2.55 & 1.78 & 0.98 & 0.65 & 0.91 & 0.58 & \textbf{0.33} & \textbf{0.26} & 1.19 & 0.82 \\
\textbf{Ours} & \textbf{2.33} & \textbf{1.70} & \textbf{0.97} & \textbf{0.66} & \textbf{0.81} & \textbf{0.55} & 0.35 & 0.27 & \textbf{1.12} & \textbf{0.80} \\
\hline
\end{tabular}
\label{tab:rgbt_scenes}
\end{table}


\subsection{RGB-NIR}

\noindent\textbf{RGB-NIR-Stereo} NIR captures the near-infrared spectrum, which is close to the visible range, and hence we compute regular image quality metrics, including PSNR, SSIM, and LPIPS, against groundtruth. Tab.~\ref{tab:rgb-nir} shows the comparison, where our method consistently achieves better scores across all the metrics, validating our pipeline. 
We show a visual comparison in Fig.~\ref{fig:NIR_results}, where our method produces better scene structure.
In Fig.~\ref{fig:temporal_consistency_rgbnir}, the lower panel, we show a visual comparison to a RGB$\to$NIR image generation method, PixNext~\cite{jin2025pix2next}. The method shows better generation consistency due to affinity to the visual spectrum, which makes image translation easier. But it still fails to produce the correct intensity compared to the groundtruth. This is also reflected in Tab.~\ref{tab:rgb-nir} with lower image quality scores.

\begin{table}[t]
\centering
\small
\setlength{\tabcolsep}{4pt}
\caption{\textbf{Results on RGB-NIR-Stereo}. Image quality metrics are shown. All methods are run with 3DGS. }
\begin{tabular}{|l|c|c|c|}
\hline
\rowcolor{gray!10}
\textbf{Method} & \textbf{PSNR}$\uparrow$ & \textbf{SSIM}$\uparrow$ & \textbf{LPIPS}$\downarrow$\\
\hline
PixNext~\cite{jin2025pix2next} & 11.283 & 0.441 & 0.452 \\
XoFTR~\cite{tuzcuouglu2024xoftr} & 14.846 & 0.321 & 0.486 \\
LightGlue~\cite{lindenberger2023lightglue} & 16.253 & 0.377 & 0.455  \\
LoFTR~\cite{sun2021loftr} & 20.179 & 0.551 & 0.356 \\
MINIMA~\cite{ren2025minima} & 20.392 & 0.568 & 0.360 \\
Ours & \textbf{21.152} & \textbf{0.581} & \textbf{0.344} \\
\hline
\end{tabular}
\label{tab:rgb-nir}
\end{table}

\noindent\textbf{Ablation Studies}. We show ablation studies in Tab.~\ref{tab:ablation}, where we gradually remove each component and show the results, including 3DGS, self-matching and filtering, DySPN confidence, multi-level thresholds, and area sampling. From the table, each proposed component contributes to the final results. By integrating the image matching confidence into densification, the PSNR increases by 1dB. The self-matching and filtering help filter out poorly generated patches and condition on more robust points for re-densification, which accounts for 0.8dB improvement. 3DGS improves the performance and gets better view synthesis.

\begin{table}[t]
\centering
\small
\setlength{\tabcolsep}{4pt}
\caption{\textbf{Ablation Study.} We ablate each component in turn from the pipeline and report the average scores on RGB-NIR-Stereo.}
\vspace{-2pt}
\begin{tabular}{|l|c|c|c|}
\hline
\rowcolor{gray!10}
\textbf{Method} & \textbf{PSNR}$\uparrow$ & \textbf{SSIM}$\uparrow$ & \textbf{LPIPS}$\downarrow$\\
\hline

Ours (final) & 21.152 & 0.581 & 0.344 \\ 
-3DGS & 21.042 & 0.597 & 0.378 \\
-Self-matching \& filtering & 20.235 & 0.522 & 0.386 \\
-DySPN confidence & 19.621 & 0.508 & 0.396 \\
-Multi-level thresholds & 19.215 & 0.495 & 0.420 \\
-Area sampling & 16.454 & 0.408 & 0.467 \\
\hline
\end{tabular}
\label{tab:ablation}
\end{table}

\noindent\textbf{Pre-GS Comparison}.We compare results without the final Gaussian Splatting, where we further discard the requirements for RGB camera poses and intrinsics from COLMAP.
Tab.~\ref{tab:preGS_comp} shows the comparison, where all the methods in comparison are run without 3DGS. Our method still performs best. Notably, even without the GS-stage, we attain a mean PSNR of 21.042 (Tab.~\ref{tab:ablation}, also averaged from 5 sequences in Tab.~\ref{tab:preGS_comp}), which is still better than all the methods run with 3DGS in Tab.~\ref{tab:rgb-nir}. The results show that our sampling and CADF strategy is successful.

\begin{table}[t]
\centering
\scriptsize
\setlength{\tabcolsep}{1pt}
\caption{\textbf{Comparison without 3DGS for all methods}. PSNR$\uparrow$ on each sequence is shown on RGB-NIR-Stereo.}
\vspace{-5pt}
\begin{tabular}{|l|p{1.0cm}|p{1.0cm}|p{1.0cm}|p{1.0cm}|p{1.0cm}|}
\hline
\rowcolor{gray!10}
\textbf{Method} & \textbf{09-28-16-48-17/1} & \textbf{09-28-16-48-17/2} & \textbf{09-28-17-06-04/1}& \textbf{09-28-17-06-04/2} & \textbf{10-26-15-53-03}\\
\hline
PixNext~\cite{jin2025pix2next} & 11.19 & 11.99 & 12.81 & 12.22 & 8.65 \\
XoFTR~\cite{tuzcuouglu2024xoftr} & 14.08 & 14.01 & 14.72 & 16.57 & 16.02 \\
LightGlue~\cite{lindenberger2023lightglue} & 14.03 & 14.10 & 14.61 & 16.67 & 16.07  \\
LoFTR~\cite{sun2021loftr} & 19.58 & 17.15 & 18.54 & 19.19 & 16.02\\
MINIMA~\cite{ren2025minima} & 21.16 & 18.80 & 19.67 & 21.76 & 19.70\\
Ours & \textbf{21.47} & \textbf{20.27} & \textbf{20.24} & \textbf{22.67} & \textbf{20.56}\\
\hline
\end{tabular}
\vspace{-2pt}
\label{tab:preGS_comp}
\end{table}

\subsection{RGB-SAR}

We conduct experiments on SAR view synthesis. We segment three DDHR-HK SAR and satellite images into overlapping 512x512 patches, and perform matching and densification on the patches. We do not perform 3DGS for all the methods in comparison and show image quality metrics against groundtruth.
From Tab.~\ref{tab:SAR_results}, ours also scores the best in all metrics, demonstrating our superiority in this challenging task, where SAR signals are challenging for cross-modal matching. 
See Supplementary for more results, visualization including SAR's, and other experiments, including feed-forward 3D reconstruction models and discussion on depth-reprojection.



\begin{table}[t]
\centering
\small
\setlength{\tabcolsep}{4pt}
\caption{\textbf{Comparison DDHR-HK SAR}. We show image quality metrics against groundtruth.}
\vspace{-4pt}
\begin{tabular}{|l|c|c|c|}
\hline
\rowcolor{gray!10}
\textbf{Method} & \textbf{PSNR}$\uparrow$ & \textbf{SSIM}$\uparrow$ & \textbf{LPIPS}$\downarrow$\\
\hline
XoFTR~\cite{tuzcuouglu2024xoftr} & 13.552 & 0.168 & 0.575 \\
LoFTR~\cite{sun2021loftr} & 13.462 & 0.164 & 0.615 \\
LightGlue~\cite{lindenberger2023lightglue} & 13.084 & 0.161 & 0.650 \\
MINIMA~\cite{ren2025minima} & 14.849 & 0.229 & 0.377 \\
Ours & \textbf{17.102} & \textbf{0.302} & \textbf{0.339} \\
\hline
\end{tabular}
\label{tab:SAR_results}
\end{table}

\section{Conclusion}
We present the first study of cross-sensor view synthesis across different types of modalities--- studying a practical, fundamental, yet widely overlooked problem in cross-modal learning on data alignment. 
We propose a novel and scalable framework 
in a match-densify-consolidate fashion 
to fulfill the goal, including integrating image matching and cross-modal densification together, using a carefully designed confidence guidance mechanism, a self-matching approach to further enhance quality, and consolidating cross-modalities in a unified 3D field. 
Extensive experiments on diverse sensor datasets validate our strategy. We expect the work to relieve the burden of sensor calibration and advance the popularity of cross-sensor learning in computer vision.

\noindent\textbf{Limitations}. First, the work focuses on static scenes and does not address dynamic objects, which is an issue in 3D consolidation and is still open for 3DGS research. Further, sensors such as thermal cameras are usually noisy and low-resolution~\cite{barral2024fixed}. Poor data quality can affect results and require in-domain denoising or enhancement on the signal level.
Still, we rely on image matching and cannot solve extremely homogeneous cases without effective descriptors, a general bottleneck for matching-based solutions.



{
    \small
    \bibliographystyle{ieeenat_fullname}
    \bibliography{main}

@String(CVPR= {IEEE Conf. Comput. Vis. Pattern Recog.})

@String(ICCV= {Int. Conf. Comput. Vis.})

@String(BMVC= {Brit. Mach. Vis. Conf.})

@String(ICASSP=	{ICASSP})

@String(AAAI = {AAAI})

@String(CVPR  = {CVPR})

@String(ICCV  = {ICCV})

@String(BMVC  =	{BMVC})

@article{choi2018kaist,
  title={KAIST multi-spectral day/night data set for autonomous and assisted driving},
  author={Choi, Yukyung and Kim, Namil and Hwang, Soonmin and Park, Kibaek and Yoon, Jae Shin and An, Kyounghwan and Kweon, In So},
  journal={IEEE Transactions on Intelligent Transportation Systems},
  volume={19},
  number={3},
  pages={934--948},
  year={2018},
  publisher={IEEE}
}

@article{li2023emergent,
  title={Emergent visual sensors for autonomous vehicles},
  author={Li, You and Moreau, Julien and Ibanez-Guzman, Javier},
  journal={IEEE Transactions on Intelligent Transportation Systems},
  volume={24},
  number={5},
  pages={4716--4737},
  year={2023},
  publisher={IEEE}
}

@article{bustos2023systematic,
  title={A systematic literature review on object detection using near infrared and thermal images},
  author={Bustos, Nicolas and Mashhadi, Mehrsa and Lai-Yuen, Susana K and Sarkar, Sudeep and Das, Tapas K},
  journal={Neurocomputing},
  volume={560},
  pages={126804},
  year={2023},
  publisher={Elsevier}
}

@article{kong2021direct,
  title={Direct near-infrared-depth visual SLAM with active lighting},
  author={Kong, Da and Zhang, Yu and Dai, Weichen},
  journal={IEEE robotics and Automation Letters},
  volume={6},
  number={4},
  pages={7057--7064},
  year={2021},
  publisher={IEEE}
}

@inproceedings{munir2021sstn,
  title={Sstn: Self-supervised domain adaptation thermal object detection for autonomous driving},
  author={Munir, Farzeen and Azam, Shoaib and Jeon, Moongu},
  booktitle={2021 IEEE/RSJ international conference on intelligent robots and systems (IROS)},
  pages={206--213},
  year={2021},
  organization={IEEE}
}

@inproceedings{franchi2024infraparis,
  title={Infraparis: A multi-modal and multi-task autonomous driving dataset},
  author={Franchi, Gianni and Hariat, Marwane and Yu, Xuanlong and Belkhir, Nacim and Manzanera, Antoine and Filliat, David},
  booktitle={Proceedings of the IEEE/CVF Winter Conference on Applications of Computer Vision},
  pages={2973--2983},
  year={2024}
}

@article{munir2022exploring,
  title={Exploring thermal images for object detection in underexposure regions for autonomous driving},
  author={Munir, Farzeen and Azam, Shoaib and Rafique, Muhammd Aasim and Sheri, Ahmad Muqeem and Jeon, Moongu and Pedrycz, Witold},
  journal={Applied soft computing},
  volume={121},
  pages={108793},
  year={2022},
  publisher={Elsevier}
}

@article{li2025improving,
  title={Improving RGB-Thermal Semantic Scene Understanding with Synthetic Data Augmentation for Autonomous Driving},
  author={Li, Haotian and Chu, Henry K and Sun, Yuxiang},
  journal={IEEE Robotics and Automation Letters},
  year={2025},
  publisher={IEEE}
}

@inproceedings{jadin2014gas,
  title={Gas leakage detection using thermal imaging technique},
  author={Jadin, Mohd Shawal and Ghazali, Kamarul Hawari},
  booktitle={2014 UKSim-AMSS 16th International Conference on Computer Modelling and Simulation},
  pages={302--306},
  year={2014},
  organization={IEEE}
}

@inproceedings{shin2023deep,
  title={Deep depth estimation from thermal image},
  author={Shin, Ukcheol and Park, Jinsun and Kweon, In So},
  booktitle={Proceedings of the IEEE/CVF Conference on Computer Vision and Pattern Recognition},
  pages={1043--1053},
  year={2023}
}

@inproceedings{fan2024generalizable,
  title={Generalizable Thermal-based Depth Estimation via Pre-trained Visual Foundation Model},
  author={Fan, Ruoyu and Zhao, Wang and Lin, Matthieu and Wang, Qi and Liu, Yong-Jin and Wang, Wenping},
  booktitle={2024 IEEE International Conference on Robotics and Automation (ICRA)},
  pages={14614--14621},
  year={2024},
  organization={IEEE}
}

@article{zhao2025unveiling,
  title={Unveiling the potential of segment anything model 2 for rgb-thermal semantic segmentation with language guidance},
  author={Zhao, Jiayi and Teng, Fei and Luo, Kai and Zhao, Guoqiang and Li, Zhiyong and Zheng, Xu and Yang, Kailun},
  journal={IEEE/RSJ international conference on intelligent robots and systems (IROS)},
  year={2025}
}

@article{paranjape2025f,
  title={F-ViTA: Foundation Model Guided Visible to Thermal Translation},
  author={Paranjape, Jay N and de Melo, Celso and Patel, Vishal M},
  journal={arXiv preprint arXiv:2504.02801},
  year={2025}
}

@inproceedings{ren2025minima,
  title={Minima: Modality invariant image matching},
  author={Ren, Jiangwei and Jiang, Xingyu and Li, Zizhuo and Liang, Dingkang and Zhou, Xin and Bai, Xiang},
  booktitle={Proceedings of the Computer Vision and Pattern Recognition Conference},
  pages={23059--23068},
  year={2025}
}

@article{yang2025distillmatch,
  title={DistillMatch: Leveraging Knowledge Distillation from Vision Foundation Model for Multimodal Image Matching},
  author={Yang, Meng and Fan, Fan and Li, Zizhuo and Deng, Songchu and Ma, Yong and Ma, Jiayi},
  journal={arXiv preprint arXiv:2509.16017},
  year={2025}
}

@inproceedings{tuzcuouglu2024xoftr,
  title={Xoftr: Cross-modal feature matching transformer},
  author={Tuzcuo{\u{g}}lu, {\"O}nder and K{\"o}ksal, Aybora and Sofu, Bu{\u{g}}ra and Kalkan, Sinan and Alatan, A Aydin},
  booktitle={Proceedings of the IEEE/CVF conference on computer vision and pattern recognition Workshops},
  pages={4275--4286},
  year={2024}
}

@inproceedings{mirlach2025r,
  title={R-livit: A lidar-visual-thermal dataset enabling vulnerable road user focused roadside perception},
  author={Mirlach, Jonas and Wan, Lei and Wiedholz, Andreas and Keen, Hannan Ejaz and Eich, Andreas},
  booktitle={ICCV},
  year={2025}
}

@inproceedings{palladin2024samfusion,
  title={Samfusion: Sensor-adaptive multimodal fusion for 3d object detection in adverse weather},
  author={Palladin, Edoardo and Dietze, Roland and Narayanan, Praveen and Bijelic, Mario and Heide, Felix},
  booktitle={European Conference on Computer Vision},
  pages={484--503},
  year={2024},
  organization={Springer}
}

@inproceedings{bijelic2020seeing,
  title={Seeing through fog without seeing fog: Deep multimodal sensor fusion in unseen adverse weather},
  author={Bijelic, Mario and Gruber, Tobias and Mannan, Fahim and Kraus, Florian and Ritter, Werner and Dietmayer, Klaus and Heide, Felix},
  booktitle={Proceedings of the IEEE/CVF Conference on Computer Vision and Pattern Recognition},
  pages={11682--11692},
  year={2020}
}

@inproceedings{shaik2024idd,
  title={Idd-aw: A benchmark for safe and robust segmentation of drive scenes in unstructured traffic and adverse weather},
  author={Shaik, Furqan Ahmed and Reddy, Abhishek and Billa, Nikhil Reddy and Chaudhary, Kunal and Manchanda, Sunny and Varma, Girish},
  booktitle={Proceedings of the IEEE/CVF Winter Conference on Applications of Computer Vision},
  pages={4614--4623},
  year={2024}
}

@inproceedings{kim2024exploiting,
  title={Exploiting cross-modal cost volume for multi-sensor depth estimation},
  author={Kim, Janghyun and Shin, Ukcheol and Heo, Seokyong and Park, Jinsun},
  booktitle={Proceedings of the Asian Conference on Computer Vision},
  pages={1420--1436},
  year={2024}
}

@inproceedings{jin2022darkvisionnet,
  title={Darkvisionnet: Low-light imaging via rgb-nir fusion with deep inconsistency prior},
  author={Jin, Shuangping and Yu, Bingbing and Jing, Minhao and Zhou, Yi and Liang, Jiajun and Ji, Renhe},
  booktitle={Proceedings of the AAAI Conference on Artificial Intelligence},
  volume={36},
  number={1},
  pages={1104--1112},
  year={2022}
}

@article{sun2019rtfnet,
  title={RTFNet: RGB-thermal fusion network for semantic segmentation of urban scenes},
  author={Sun, Yuxiang and Zuo, Weixun and Liu, Ming},
  journal={IEEE Robotics and Automation Letters},
  volume={4},
  number={3},
  pages={2576--2583},
  year={2019},
  publisher={IEEE}
}

@inproceedings{kim2025pixelnir,
  title={Pixel-aligned RGB-NIR Stereo Imaging and Dataset for Robot Vision},
  author={Kim, Jinnyeong and Baek, Seung-Hwan},
  booktitle={Proceedings of the Computer Vision and Pattern Recognition Conference},
  pages={11482--11492},
  year={2025}
}

@article{liang2023explicit,
  title={Explicit attention-enhanced fusion for RGB-thermal perception tasks},
  author={Liang, Mingjian and Hu, Junjie and Bao, Chenyu and Feng, Hua and Deng, Fuqin and Lam, Tin Lun},
  journal={IEEE Robotics and Automation Letters},
  volume={8},
  number={7},
  pages={4060--4067},
  year={2023},
  publisher={IEEE}
}

@inproceedings{liu2023multi,
  title={Multi-interactive feature learning and a full-time multi-modality benchmark for image fusion and segmentation},
  author={Liu, Jinyuan and Liu, Zhu and Wu, Guanyao and Ma, Long and Liu, Risheng and Zhong, Wei and Luo, Zhongxuan and Fan, Xin},
  booktitle={Proceedings of the IEEE/CVF international conference on computer vision},
  pages={8115--8124},
  year={2023}
}

@article{zhang2023cmx,
  title={CMX: Cross-modal fusion for RGB-X semantic segmentation with transformers},
  author={Zhang, Jiaming and Liu, Huayao and Yang, Kailun and Hu, Xinxin and Liu, Ruiping and Stiefelhagen, Rainer},
  journal={IEEE Transactions on intelligent transportation systems},
  volume={24},
  number={12},
  pages={14679--14694},
  year={2023},
  publisher={IEEE}
}

@article{tang2024divide,
  title={Divide-and-conquer: Confluent triple-flow network for RGB-T salient object detection},
  author={Tang, Hao and Li, Zechao and Zhang, Dong and He, Shengfeng and Tang, Jinhui},
  journal={IEEE Transactions on Pattern Analysis and Machine Intelligence},
  year={2024},
  publisher={IEEE}
}

@article{zhou2021gmnet,
  title={GMNet: Graded-feature multilabel-learning network for RGB-thermal urban scene semantic segmentation},
  author={Zhou, Wujie and Liu, Jinfu and Lei, Jingsheng and Yu, Lu and Hwang, Jenq-Neng},
  journal={IEEE Transactions on Image Processing},
  volume={30},
  pages={7790--7802},
  year={2021},
  publisher={IEEE}
}

@inproceedings{wan2025sigma,
  title={Sigma: Siamese mamba network for multi-modal semantic segmentation},
  author={Wan, Zifu and Zhang, Pingping and Wang, Yuhao and Yong, Silong and Stepputtis, Simon and Sycara, Katia and Xie, Yaqi},
  booktitle={2025 IEEE/CVF Winter Conference on Applications of Computer Vision (WACV)},
  pages={1734--1744},
  year={2025},
  organization={IEEE}
}

@article{liu2024infrared,
  title={Infrared and visible image fusion: From data compatibility to task adaption},
  author={Liu, Jinyuan and Wu, Guanyao and Liu, Zhu and Wang, Di and Jiang, Zhiying and Ma, Long and Zhong, Wei and Fan, Xin},
  journal={IEEE Transactions on Pattern Analysis and Machine Intelligence},
  year={2024},
  publisher={IEEE}
}

@inproceedings{zheng2024learning,
  title={Learning modality-agnostic representation for semantic segmentation from any modalities},
  author={Zheng, Xu and Lyu, Yuanhuiyi and Wang, Lin},
  booktitle={European Conference on Computer Vision},
  pages={146--165},
  year={2024},
  organization={Springer}
}

@inproceedings{zhang2021abmdrnet,
  title={ABMDRNet: Adaptive-weighted bi-directional modality difference reduction network for RGB-T semantic segmentation},
  author={Zhang, Qiang and Zhao, Shenlu and Luo, Yongjiang and Zhang, Dingwen and Huang, Nianchang and Han, Jungong},
  booktitle={Proceedings of the IEEE/CVF Conference on Computer Vision and Pattern Recognition},
  pages={2633--2642},
  year={2021}
}

@inproceedings{deng2021feanet,
  title={FEANet: Feature-enhanced attention network for RGB-thermal real-time semantic segmentation},
  author={Deng, Fuqin and Feng, Hua and Liang, Mingjian and Wang, Hongmin and Yang, Yong and Gao, Yuan and Chen, Junfeng and Hu, Junjie and Guo, Xiyue and Lam, Tin Lun},
  booktitle={2021 IEEE/RSJ international conference on intelligent robots and systems (IROS)},
  pages={4467--4473},
  year={2021},
  organization={IEEE}
}

@inproceedings{ji2023multispectral,
  title={Multispectral video semantic segmentation: A benchmark dataset and baseline},
  author={Ji, Wei and Li, Jingjing and Bian, Cheng and Zhou, Zongwei and Zhao, Jiaying and Yuille, Alan L and Cheng, Li},
  booktitle={Proceedings of the IEEE/CVF Conference on Computer Vision and Pattern Recognition},
  pages={1094--1104},
  year={2023}
}

@article{feng2023cekd,
  title={CEKD: Cross-modal edge-privileged knowledge distillation for semantic scene understanding using only thermal images},
  author={Feng, Zhen and Guo, Yanning and Sun, Yuxiang},
  journal={IEEE Robotics and Automation Letters},
  volume={8},
  number={4},
  pages={2205--2212},
  year={2023},
  publisher={IEEE}
}

@article{xu2021multimodal,
  title={Multimodal cross-layer bilinear pooling for RGBT tracking},
  author={Xu, Qin and Mei, Yiming and Liu, Jinpei and Li, Chenglong},
  journal={IEEE Transactions on Multimedia},
  volume={24},
  pages={567--580},
  year={2021},
  publisher={IEEE}
}

@inproceedings{deevi2024rgb,
  title={Rgb-x object detection via scene-specific fusion modules},
  author={Deevi, Sri Aditya and Lee, Connor and Gan, Lu and Nagesh, Sushruth and Pandey, Gaurav and Chung, Soon-Jo},
  booktitle={Proceedings of the IEEE/CVF Winter Conference on Applications of Computer Vision},
  pages={7366--7375},
  year={2024}
}

@inproceedings{li2025pseudo,
  title={Pseudo Visible Feature Fine-Grained Fusion for Thermal Object Detection},
  author={Li, Ting and Ye, Mao and Wu, Tianwen and Li, Nianxin and Li, Shuaifeng and Tang, Song and Ji, Luping},
  booktitle={Proceedings of the Computer Vision and Pattern Recognition Conference},
  pages={6710--6719},
  year={2025}
}

@article{xiao2025thermalgen,
  title={ThermalGen: Style-Disentangled Flow-Based Generative Models for RGB-to-Thermal Image Translation},
  author={Xiao, Jiuhong and Nayak, Roshan and Zhang, Ning and Tortei, Daniel and Loianno, Giuseppe},
  journal={arXiv preprint arXiv:2509.24878},
  year={2025}
}

@inproceedings{abbott2020unsupervised,
  title={Unsupervised object detection via LWIR/RGB translation},
  author={Abbott, Rachael and Robertson, Neil M and del Rincon, Jesus Martinez and Connor, Barry},
  booktitle={Proceedings of the IEEE/CVF Conference on Computer Vision and Pattern Recognition Workshops},
  pages={90--91},
  year={2020}
}

@inproceedings{frigo2022doodlenet,
  title={DooDLeNet: Double DeepLab enhanced feature fusion for thermal-color semantic segmentation},
  author={Frigo, Oriel and Martin-Gaffe, Lucien and Wacongne, Catherine},
  booktitle={Proceedings of the IEEE/CVF Conference on Computer Vision and Pattern Recognition},
  pages={3021--3029},
  year={2022}
}

@inproceedings{berg2018generating,
  title={Generating visible spectrum images from thermal infrared},
  author={Berg, Amanda and Ahlberg, Jorgen and Felsberg, Michael},
  booktitle={Proceedings of the IEEE Conference on Computer Vision and Pattern Recognition Workshops},
  pages={1143--1152},
  year={2018}
}

@inproceedings{madan2024rabbit,
  title={Rabbit: A robot-assisted bed bathing system with multimodal perception and integrated compliance},
  author={Madan, Rishabh and Valdez, Skyler and Kim, David and Fang, Sujie and Zhong, Luoyan and Virtue, Diego T and Bhattacharjee, Tapomayukh},
  booktitle={Proceedings of the 2024 ACM/IEEE international conference on human-robot interaction},
  pages={472--481},
  year={2024}
}

@inproceedings{li2020challenge,
  title={Challenge-aware RGBT tracking},
  author={Li, Chenglong and Liu, Lei and Lu, Andong and Ji, Qing and Tang, Jin},
  booktitle={European conference on computer vision},
  pages={222--237},
  year={2020},
  organization={Springer}
}

@inproceedings{hong2024onetracker,
  title={Onetracker: Unifying visual object tracking with foundation models and efficient tuning},
  author={Hong, Lingyi and Yan, Shilin and Zhang, Renrui and Li, Wanyun and Zhou, Xinyu and Guo, Pinxue and Jiang, Kaixun and Chen, Yiting and Li, Jinglun and Chen, Zhaoyu and others},
  booktitle={Proceedings of the IEEE/CVF conference on computer vision and pattern recognition},
  pages={19079--19091},
  year={2024}
}

@inproceedings{hou2024sdstrack,
  title={Sdstrack: Self-distillation symmetric adapter learning for multi-modal visual object tracking},
  author={Hou, Xiaojun and Xing, Jiazheng and Qian, Yijie and Guo, Yaowei and Xin, Shuo and Chen, Junhao and Tang, Kai and Wang, Mengmeng and Jiang, Zhengkai and Liu, Liang and others},
  booktitle={Proceedings of the IEEE/CVF Conference on Computer Vision and Pattern Recognition},
  pages={26551--26561},
  year={2024}
}

@inproceedings{zhang2022visible,
  title={Visible-thermal UAV tracking: A large-scale benchmark and new baseline},
  author={Zhang, Pengyu and Zhao, Jie and Wang, Dong and Lu, Huchuan and Ruan, Xiang},
  booktitle={Proceedings of the IEEE/CVF Conference on Computer Vision and Pattern Recognition},
  pages={8886--8895},
  year={2022}
}

@inproceedings{hui2023bridging,
  title={Bridging search region interaction with template for rgb-t tracking},
  author={Hui, Tianrui and Xun, Zizheng and Peng, Fengguang and Huang, Junshi and Wei, Xiaoming and Wei, Xiaolin and Dai, Jiao and Han, Jizhong and Liu, Si},
  booktitle={Proceedings of the IEEE/CVF Conference on Computer Vision and Pattern Recognition},
  pages={13630--13639},
  year={2023}
}

@inproceedings{zhang2023efficient,
  title={Efficient rgb-t tracking via cross-modality distillation},
  author={Zhang, Tianlu and Guo, Hongyuan and Jiao, Qiang and Zhang, Qiang and Han, Jungong},
  booktitle={Proceedings of the IEEE/CVF conference on computer vision and pattern recognition},
  pages={5404--5413},
  year={2023}
}

@inproceedings{aditya2024thermal,
  title={Thermal voyager: A comparative study of rgb and thermal cameras for night-time autonomous navigation},
  author={Aditya, NG and Dhruval, PB and others},
  booktitle={2024 IEEE International Conference on Robotics and Automation (ICRA)},
  pages={14116--14122},
  year={2024},
  organization={IEEE}
}

@article{jiang2022thermal,
  title={Thermal-inertial SLAM for the environments with challenging illumination},
  author={Jiang, Jiajun and Chen, Xingxin and Dai, Weichen and Gao, Zelin and Zhang, Yu},
  journal={IEEE Robotics and Automation Letters},
  volume={7},
  number={4},
  pages={8767--8774},
  year={2022},
  publisher={IEEE}
}

@inproceedings{brown2005multi,
  title={Multi-image matching using multi-scale oriented patches},
  author={Brown, Matthew and Szeliski, Richard and Winder, Simon},
  booktitle={2005 IEEE Computer Society Conference on Computer Vision and Pattern Recognition (CVPR'05)},
  volume={1},
  pages={510--517},
  year={2005},
  organization={IEEE}
}

@inproceedings{tola2008fast,
  title={A fast local descriptor for dense matching},
  author={Tola, Engin and Lepetit, Vincent and Fua, Pascal},
  booktitle={2008 IEEE conference on computer vision and pattern recognition},
  pages={1--8},
  year={2008},
  organization={IEEE}
}

@article{fischler1981random,
  title={Random sample consensus: a paradigm for model fitting with applications to image analysis and automated cartography},
  author={Fischler, Martin A and Bolles, Robert C},
  journal={Communications of the ACM},
  volume={24},
  number={6},
  pages={381--395},
  year={1981},
  publisher={ACM New York, NY, USA}
}

@inproceedings{dusmanu2019d2,
  title={D2-net: A trainable cnn for joint description and detection of local features},
  author={Dusmanu, Mihai and Rocco, Ignacio and Pajdla, Tomas and Pollefeys, Marc and Sivic, Josef and Torii, Akihiko and Sattler, Torsten},
  booktitle={Proceedings of the ieee/cvf conference on computer vision and pattern recognition},
  pages={8092--8101},
  year={2019}
}

@article{ono2018lf,
  title={LF-Net: Learning local features from images},
  author={Ono, Yuki and Trulls, Eduard and Fua, Pascal and Yi, Kwang Moo},
  journal={Advances in neural information processing systems},
  volume={31},
  year={2018}
}

@article{tyszkiewicz2020disk,
  title={Disk: Learning local features with policy gradient},
  author={Tyszkiewicz, Micha{\l} and Fua, Pascal and Trulls, Eduard},
  journal={Advances in neural information processing systems},
  volume={33},
  pages={14254--14265},
  year={2020}
}

@inproceedings{wang2021p2,
  title={P2-net: Joint description and detection of local features for pixel and point matching},
  author={Wang, Bing and Chen, Changhao and Cui, Zhaopeng and Qin, Jie and Lu, Chris Xiaoxuan and Yu, Zhengdi and Zhao, Peijun and Dong, Zhen and Zhu, Fan and Trigoni, Niki and others},
  booktitle={Proceedings of the IEEE/CVF International Conference on Computer Vision},
  pages={16004--16013},
  year={2021}
}

@inproceedings{sarlin2020superglue,
  title={Superglue: Learning feature matching with graph neural networks},
  author={Sarlin, Paul-Edouard and DeTone, Daniel and Malisiewicz, Tomasz and Rabinovich, Andrew},
  booktitle={Proceedings of the IEEE/CVF conference on computer vision and pattern recognition},
  pages={4938--4947},
  year={2020}
}

@inproceedings{sun2021loftr,
  title={LoFTR: Detector-free local feature matching with transformers},
  author={Sun, Jiaming and Shen, Zehong and Wang, Yuang and Bao, Hujun and Zhou, Xiaowei},
  booktitle={Proceedings of the IEEE/CVF conference on computer vision and pattern recognition},
  pages={8922--8931},
  year={2021}
}

@inproceedings{lindenberger2023lightglue,
  title={Lightglue: Local feature matching at light speed},
  author={Lindenberger, Philipp and Sarlin, Paul-Edouard and Pollefeys, Marc},
  booktitle={Proceedings of the IEEE/CVF international conference on computer vision},
  pages={17627--17638},
  year={2023}
}

@inproceedings{potje2024xfeat,
  title={Xfeat: Accelerated features for lightweight image matching},
  author={Potje, Guilherme and Cadar, Felipe and Araujo, Andr{\'e} and Martins, Renato and Nascimento, Erickson R},
  booktitle={Proceedings of the IEEE/CVF Conference on Computer Vision and Pattern Recognition},
  pages={2682--2691},
  year={2024}
}

@InProceedings{Edstedt_2024_CVPR,
    author    = {Edstedt, Johan and Sun, Qiyu and B\"okman, Georg and Wadenb\"ack, M\r{a}rten and Felsberg, Michael},
    title     = {RoMa: Robust Dense Feature Matching},
    booktitle = {Proceedings of the IEEE/CVF Conference on Computer Vision and Pattern Recognition (CVPR)},
    month     = {June},
    year      = {2024},
    pages     = {19790-19800}
}

@InProceedings{Jiang_2024_CVPR,
    author    = {Jiang, Hanwen and Karpur, Arjun and Cao, Bingyi and Huang, Qixing and Araujo, Andr\'e},
    title     = {OmniGlue: Generalizable Feature Matching with Foundation Model Guidance},
    booktitle = {Proceedings of the IEEE/CVF Conference on Computer Vision and Pattern Recognition (CVPR)},
    month     = {June},
    year      = {2024},
    pages     = {19865-19875}
}

@article{ni2024eto,
  title={Eto: Efficient transformer-based local feature matching by organizing multiple homography hypotheses},
  author={Ni, Junjie and Zhang, Guofeng and Li, Guanglin and Li, Yijin and Liu, Xinyang and Huang, Zhaoyang and Bao, Hujun},
  journal={Advances in Neural Information Processing Systems},
  volume={37},
  pages={60260--60274},
  year={2024}
}

@inproceedings{wang2022matchformer,
  title={Matchformer: Interleaving attention in transformers for feature matching},
  author={Wang, Qing and Zhang, Jiaming and Yang, Kailun and Peng, Kunyu and Stiefelhagen, Rainer},
  booktitle={Proceedings of the Asian conference on computer vision},
  pages={2746--2762},
  year={2022}
}

@inproceedings{zhang2019deep,
  title={Deep graphical feature learning for the feature matching problem},
  author={Zhang, Zhen and Lee, Wee Sun},
  booktitle={Proceedings of the IEEE/CVF International Conference on Computer Vision},
  pages={5087--5096},
  year={2019}
}

@inproceedings{shi2022clustergnn,
  title={Clustergnn: Cluster-based coarse-to-fine graph neural network for efficient feature matching},
  author={Shi, Yan and Cai, Jun-Xiong and Shavit, Yoli and Mu, Tai-Jiang and Feng, Wensen and Zhang, Kai},
  booktitle={Proceedings of the IEEE/CVF conference on computer vision and pattern recognition},
  pages={12517--12526},
  year={2022}
}

@inproceedings{wang2024dust3r,
  title={Dust3r: Geometric 3d vision made easy},
  author={Wang, Shuzhe and Leroy, Vincent and Cabon, Yohann and Chidlovskii, Boris and Revaud, Jerome},
  booktitle={Proceedings of the IEEE/CVF Conference on Computer Vision and Pattern Recognition},
  pages={20697--20709},
  year={2024}
}

@inproceedings{wang2025vggt,
  title={Vggt: Visual geometry grounded transformer},
  author={Wang, Jianyuan and Chen, Minghao and Karaev, Nikita and Vedaldi, Andrea and Rupprecht, Christian and Novotny, David},
  booktitle={Proceedings of the Computer Vision and Pattern Recognition Conference},
  pages={5294--5306},
  year={2025}
}

@inproceedings{keetha2025mapanything,
  title={MapAnything: Universal feed-forward metric 3D reconstruction},
  author={Keetha, Nikhil and M{\"u}ller, Norman and Sch{\"o}nberger, Johannes and Porzi, Lorenzo and Zhang, Yuchen and Fischer, Tobias and Knapitsch, Arno and Zauss, Duncan and Weber, Ethan and Antunes, Nelson and others},
  booktitle={3DV},
  year={2026}
}

@article{deng2022redfeat,
  title={ReDFeat: Recoupling detection and description for multimodal feature learning},
  author={Deng, Yuxin and Ma, Jiayi},
  journal={IEEE Transactions on Image Processing},
  volume={32},
  pages={591--602},
  year={2022},
  publisher={IEEE}
}

@inproceedings{edstedt2023dkm,
  title={DKM: Dense kernelized feature matching for geometry estimation},
  author={Edstedt, Johan and Athanasiadis, Ioannis and Wadenb{\"a}ck, M{\aa}rten and Felsberg, Michael},
  booktitle={Proceedings of the IEEE/CVF Conference on Computer Vision and Pattern Recognition},
  pages={17765--17775},
  year={2023}
}

@inproceedings{chen2022aspanformer,
  title={Aspanformer: Detector-free image matching with adaptive span transformer},
  author={Chen, Hongkai and Luo, Zixin and Zhou, Lei and Tian, Yurun and Zhen, Mingmin and Fang, Tian and Mckinnon, David and Tsin, Yanghai and Quan, Long},
  booktitle={European conference on computer vision},
  pages={20--36},
  year={2022},
  organization={Springer}
}

@inproceedings{zuo2025omni,
  title={Omni-dc: Highly robust depth completion with multiresolution depth integration},
  author={Zuo, Yiming and Yang, Willow and Ma, Zeyu and Deng, Jia},
  booktitle={Proceedings of the IEEE/CVF International Conference on Computer Vision},
  pages={9287--9297},
  year={2025}
}

@inproceedings{zuo2024ogni,
  title={Ogni-dc: Robust depth completion with optimization-guided neural iterations},
  author={Zuo, Yiming and Deng, Jia},
  booktitle={European Conference on Computer Vision},
  pages={78--95},
  year={2024},
  organization={Springer}
}

@inproceedings{lin2022dynamic,
  title={Dynamic spatial propagation network for depth completion},
  author={Lin, Yuankai and Cheng, Tao and Zhong, Qi and Zhou, Wending and Yang, Hua},
  booktitle={Proceedings of the AAAI conference on artificial intelligence},
  volume={36},
  number={2},
  pages={1638--1646},
  year={2022}
}

@article{ren2024grounded,
  title={Grounded sam: Assembling open-world models for diverse visual tasks},
  author={Ren, Tianhe and Liu, Shilong and Zeng, Ailing and Lin, Jing and Li, Kunchang and Cao, He and Chen, Jiayu and Huang, Xinyu and Chen, Yukang and Yan, Feng and others},
  journal={arXiv preprint arXiv:2401.14159},
  year={2024}
}

@inproceedings{luthermalgaussian,
  title={ThermalGaussian: Thermal 3D Gaussian Splatting},
  author={Lu, Rongfeng and Chen, Hangyu and Zhu, Zunjie and Qin, Yuhang and Lu, Ming and Yan, Chenggang and others},
  booktitle={The Thirteenth International Conference on Learning Representations},
  year={2025}
}

@article{sa2022deepnir,
  title={deepNIR: Datasets for generating synthetic NIR images and improved fruit detection system using deep learning techniques},
  author={Sa, Inkyu and Lim, Jong Yoon and Ahn, Ho Seok and MacDonald, Bruce},
  journal={Sensors},
  volume={22},
  number={13},
  pages={4721},
  year={2022},
  publisher={MDPI}
}

@article{cardona2024dataset,
  title={Dataset of Sentinel-1 SAR and Sentinel-2 RGB-NDVI imagery},
  author={Cardona-Mesa, Ahmed Alejandro and V{\'a}squez-Salazar, Rub{\'e}n Dar{\'\i}o and G{\'o}mez, Luis and Travieso-Gonz{\'a}lez, Carlos M and Garavito-Gonz{\'a}lez, Andr{\'e}s F and V{\'a}squez-Cano, Esteban and D{\'\i}az-Paz, Jean Pierre},
  journal={Data in Brief},
  volume={57},
  pages={111160},
  year={2024},
  publisher={Elsevier}
}

@article{xiang2020automatic,
  title={Automatic registration of optical and SAR images via improved phase congruency model},
  author={Xiang, Yuming and Tao, Rongshu and Wang, Feng and You, Hongjian and Han, Bing},
  journal={IEEE Journal of Selected Topics in Applied Earth Observations and Remote Sensing},
  volume={13},
  pages={5847--5861},
  year={2020},
  publisher={IEEE}
}

@article{ren2022dual,
  title={A dual-stream high resolution network: Deep fusion of GF-2 and GF-3 data for land cover classification},
  author={Ren, Bo and Ma, Shibin and Hou, Biao and Hong, Danfeng and Chanussot, Jocelyn and Wang, Jianlong and Jiao, Licheng},
  journal={International Journal of Applied Earth Observation and Geoinformation},
  volume={112},
  pages={102896},
  year={2022},
  publisher={Elsevier}
}

@misc{zhao2025remotesensingdataset,
  author       = {Zhao, Xuyang and Zhao, Lijun and Li, Hongyi and Zhang, Zheng and Tang, Ping},
  title        = {A Multimodal Visible-SAR Dataset for Airport Detection in Remote Sensing Imagery},
  howpublished = {Science Data Bank, Dataset, Version 1.0},
  year         = {2025},
  note         = {Available at \url{https://doi.org/10.57760/sciencedb.j00001.01276}},
  doi          = {10.57760/sciencedb.j00001.01276}
}

@article{li2022mcanet,
  title={MCANet: A joint semantic segmentation framework of optical and SAR images for land use classification},
  author={Li, Xue and Zhang, Guo and Cui, Hao and Hou, Shasha and Wang, Shunyao and Li, Xin and Chen, Yujia and Li, Zhijiang and Zhang, Li},
  journal={International Journal of Applied Earth Observation and Geoinformation},
  volume={106},
  pages={102638},
  year={2022},
  publisher={Elsevier}
}

@inproceedings{li2023blip,
  title={Blip-2: Bootstrapping language-image pre-training with frozen image encoders and large language models},
  author={Li, Junnan and Li, Dongxu and Savarese, Silvio and Hoi, Steven},
  booktitle={International conference on machine learning},
  pages={19730--19742},
  year={2023},
  organization={PMLR}
}

@article{tschannen2025siglip,
  title={Siglip 2: Multilingual vision-language encoders with improved semantic understanding, localization, and dense features},
  author={Tschannen, Michael and Gritsenko, Alexey and Wang, Xiao and Naeem, Muhammad Ferjad and Alabdulmohsin, Ibrahim and Parthasarathy, Nikhil and Evans, Talfan and Beyer, Lucas and Xia, Ye and Mustafa, Basil and others},
  journal={arXiv preprint arXiv:2502.14786},
  year={2025}
}

@InProceedings{Han_2025_ICCV,
    author    = {Han, Zhen and Mao, Chaojie and Jiang, Zeyinzi and Pan, Yulin and Zhang, Jingfeng},
    title     = {StyleBooth: Image Style Editing with Multimodal Instruction},
    booktitle = {Proceedings of the IEEE/CVF International Conference on Computer Vision (ICCV) Workshops},
    month     = {October},
    year      = {2025},
    pages     = {1947-1957}
}

@article{jin2025pix2next,
  title={Pix2next: Leveraging vision foundation models for rgb to nir image translation},
  author={Jin, Youngwan and Park, Incheol and Song, Hanbin and Ju, Hyeongjin and Nalcakan, Yagiz and Kim, Shiho},
  journal={Technologies},
  year={2025}
}

@InProceedings{Agustsson_2017_CVPR_Workshops,
    author = {Agustsson, Eirikur and Timofte, Radu},
    title = {NTIRE 2017 Challenge on Single Image Super-Resolution: Dataset and Study},
    booktitle = {The IEEE Conference on Computer Vision and Pattern Recognition (CVPR) Workshops},
    month = {July},
    year = {2017}
}

@inproceedings{peng3d,
  title={3D Vision-Language Gaussian Splatting},
  author={Peng, Qucheng and Planche, Benjamin and Gao, Zhongpai and Zheng, Meng and Choudhuri, Anwesa and Chen, Terrence and Chen, Chen and Wu, Ziyan},
  booktitle={The Thirteenth International Conference on Learning Representations},
  year = {2025}
}

@inproceedings{katragadda2025online,
  title={Online Language Splatting},
  author={Katragadda, Saimouli and Wu, Cho-Ying and Guo, Yuliang and Huang, Xinyu and Huang, Guoquan and Ren, Liu},
  booktitle={Proceedings of the IEEE/CVF International Conference on Computer Vision (ICCV)},
  year={2025}
}

@inproceedings{li2018megadepth,
  title={Megadepth: Learning single-view depth prediction from internet photos},
  author={Li, Zhengqi and Snavely, Noah},
  booktitle={Proceedings of the IEEE conference on computer vision and pattern recognition},
  pages={2041--2050},
  year={2018}
}

@inproceedings{asim2025met3r,
  title={Met3r: Measuring multi-view consistency in generated images},
  author={Asim, Mohammad and Wewer, Christopher and Wimmer, Thomas and Schiele, Bernt and Lenssen, Jan Eric},
  booktitle={Proceedings of the Computer Vision and Pattern Recognition Conference},
  pages={6034--6044},
  year={2025}
}

@article{yang2024depth,
  title={Depth anything v2},
  author={Yang, Lihe and Kang, Bingyi and Huang, Zilong and Zhao, Zhen and Xu, Xiaogang and Feng, Jiashi and Zhao, Hengshuang},
  journal={Advances in Neural Information Processing Systems},
  volume={37},
  pages={21875--21911},
  year={2024}
}

@inproceedings{schonberger2016structure,
  title={Structure-from-motion revisited},
  author={Schonberger, Johannes L and Frahm, Jan-Michael},
  booktitle={Proceedings of the IEEE conference on computer vision and pattern recognition},
  pages={4104--4113},
  year={2016}
}

@inproceedings{barral2024fixed,
  title={Fixed pattern noise removal for multi-view single-sensor infrared camera},
  author={Barral, Arnaud and Arias, Pablo and Davy, Axel},
  booktitle={Proceedings of the IEEE/CVF Winter Conference on Applications of Computer Vision},
  pages={1669--1678},
  year={2024}
}

@InProceedings{Wang_2022_CVPR,
    author    = {Wang, Yan},
    title     = {Edge-Enhanced Feature Distillation Network for Efficient Super-Resolution},
    booktitle = {Proceedings of the IEEE/CVF Conference on Computer Vision and Pattern Recognition (CVPR) Workshops},
    month     = {June},
    year      = {2022},
    pages     = {777-785}
}

@article{kerbl20233d,
  title={3d gaussian splatting for real-time radiance field rendering.},
  author={Kerbl, Bernhard and Kopanas, Georgios and Leimk{\"u}hler, Thomas and Drettakis, George and others},
  journal={ACM Trans. Graph.},
  volume={42},
  number={4},
  pages={139--1},
  year={2023}
}

@inproceedings{yu2024mip,
  title={Mip-splatting: Alias-free 3d gaussian splatting},
  author={Yu, Zehao and Chen, Anpei and Huang, Binbin and Sattler, Torsten and Geiger, Andreas},
  booktitle={Proceedings of the IEEE/CVF conference on computer vision and pattern recognition},
  pages={19447--19456},
  year={2024}
}

@inproceedings{huang3dgeer,
  title={3DGEER: 3D Gaussian Rendering Made Exact and Efficient for Generic Cameras},
  author={Huang, Zixun and Wu, Cho-Ying and Guo, Yuliang and Huang, Xinyu and Ren, Liu},
  booktitle={The Fourteenth International Conference on Learning Representations},
  year={2026}
}

@inproceedings{wu2023inspacetype,
  title={InSpaceType: Reconsider space type in indoor monocular depth estimation},
  author={Wu, Cho-Ying and Gao, Quankai and Hsu, Chin-Cheng and Wu, Te-Lin and Chen, Jing-Wen and Neumann, Ulrich},
  booktitle={The British Machine Vision Conference (BMVC)},
  year={2024}
}

@InProceedings{Wu_2022_CVPR,
    author    = {Wu, Cho-Ying and Wang, Jialiang and Hall, Michael and Neumann, Ulrich and Su, Shuochen},
    title     = {Toward Practical Monocular Indoor Depth Estimation},
    booktitle = {Proceedings of the IEEE/CVF Conference on Computer Vision and Pattern Recognition (CVPR)},
    month     = {June},
    year      = {2022},
    pages     = {3814-3824}
}

@inproceedings{wu2024boosting,
  title={Boosting Generalizability towards Zero-Shot Cross-Dataset Single-Image Indoor Depth by Meta-Initialization},
  author={Wu, Cho-Ying and Zhong, Yiqi and Wang, Junying and Neumann, Ulrich},
  booktitle={2024 IEEE/RSJ International Conference on Intelligent Robots and Systems (IROS)},
  pages={10051--10058},
  year={2024},
  organization={IEEE}
}

@inproceedings{guo2025depth,
  title={Depth any camera: Zero-shot metric depth estimation from any camera},
  author={Guo, Yuliang and Garg, Sparsh and Miangoleh, S Mahdi H and Huang, Xinyu and Ren, Liu},
  booktitle={Proceedings of the Computer Vision and Pattern Recognition Conference},
  pages={26996--27006},
  year={2025}
}

@inproceedings{wu2021scene,
  title={Scene completeness-aware lidar depth completion for driving scenario},
  author={Wu, Cho-Ying and Neumann, Ulrich},
  booktitle={ICASSP 2021-2021 IEEE International Conference on Acoustics, Speech and Signal Processing (ICASSP)},
  pages={2490--2494},
  year={2021},
  organization={IEEE}
}
}


\end{document}